\documentclass[table]{mintreport}
\PassOptionsToPackage{table,usenames,dvipsnames}{xcolor}

\usepackage{amssymb}
\usepackage{multirow}
\usepackage{multicol}
\usepackage{marvosym}
\usepackage{longtable}
\usepackage{tabularray}
\usepackage{fontawesome5}
\usepackage{wrapfig}
\usepackage[most]{tcolorbox}
\usepackage{url}
\usepackage{float}
\usepackage{enumitem}
\usepackage{subcaption}
\usepackage{caption}
\usepackage{makecell}
\usepackage{adjustbox}
\usepackage{booktabs}
\usepackage{array}
\usepackage{amsfonts}
\usepackage{nicefrac}
\usepackage{microtype}
\usepackage{xspace}
\usepackage{amsthm}
\usepackage{amsmath}
\usepackage{bm}
\usepackage{pifont}
\usepackage{fontawesome5}
\usepackage{CJKutf8}

\definecolor{mintphone}{HTML}{D65A31}
\definecolor{githubmark}{HTML}{24292F}
\definecolor{bestcell}{HTML}{86CFA5}
\definecolor{secondcell}{HTML}{C6EAD2}
\definecolor{thirdcell}{HTML}{E8F6EC}
\definecolor{promptteal}{HTML}{2F7F78}
\definecolor{prompttealsoft}{HTML}{EAF6F3}

\RequirePackage{tgpagella}
\RequirePackage{mathpazo}
\RequirePackage{inconsolata}

\newcolumntype{C}[1]{>{\centering\arraybackslash}m{#1}}

\theoremstyle{definition}

\renewcommand{\title}[1]{\newcommand{\titlelist}{{\huge\fontfamily{optimistic}\selectfont #1}}}

\newcommand{\ourmethod}{Mint-Agent\xspace}
\newcommand{\ourcu}{Mint-Cu\xspace}
\newcommand{\ourag}{Mint-Ag\xspace}

\newtcolorbox{promptbox}{
  enhanced,
  breakable,
  colback=prompttealsoft,
  colframe=promptteal,
  boxrule=0.7pt,
  arc=2mm,
  left=3mm,
  right=3mm,
  top=2.5mm,
  bottom=2.5mm,
  before skip=0.6em,
  after skip=0.8em
}

\title{
\huge{\textcolor{mintdark}{Mint-Agent}: Introducing Finance-Native Agentic Foundation Models}
}

\author{
    Kun Wang$^{1,2}$,
    Gavin Zhang$^{1}$,
    Yaze Geng$^{1}$,
    Lei Tang$^{1}$,
    Yaoyang Yi$^{1}$,
    Zonghan Wu$^{1,3}$,
    Yifan Hu$^{4}$,
    Qingsong Wen$^{3,}$,
    Yilei Shao$^{1,3,}\textsuperscript{\faEnvelope}$
    \\[0.5em]
    \small
    $^{1}$Mint Agent Team\;\;
    $^{2}$Nanyang Technological University\;\\
    $^{3}$Shanghai AI-Finance School, East China Normal University\;\;
    $^{4}$Tsinghua University\;\;
    \\[0.5em]
    \small \faEnvelope\;
    wang.kun@minttruth.com,
    yileishao@sem.ecnu.edu.cn
}

\abstract{
Financial agents must do more than recall domain knowledge: they must be both \textbf{reliable}, executing precise operations over grounded evidence, and \textbf{executive}, sustaining long-horizon research whose conclusions remain auditable. We present \ourmethod, a family of finance-native agentic models designed around these two scales of financial intelligence. \ourmethod is built upon three pillars: \textbf{data}, \textbf{harness}, and \textbf{algorithm}. Our data engine constructs clean, specialized tasks for atomic financial capabilities and long-horizon agentic execution from real-world financial sources. MintHarness enables stable interaction with open-ended environments and maintains auditable evidence trails across extended research trajectories. Our training recipe combines supervised fine-tuning (SFT), on-policy distillation (OPD), and reinforcement learning with verifiable rewards (RLVR) to develop separate financial reasoning and agentic execution experts, which are then unified through model merging and multi-teacher on-policy distillation into compact, general-purpose financial agents. This pipeline yields two flagship models, \ourcu{} (9B) and \ourag{} (27B).
Across professional financial benchmarks, our models demonstrate two defining strengths: \ding{182} \textbf{Reliability}: \ourag{} achieves 98.33\% on RFC-Bench, surpassing GPT-5.6-Sol and Claude Opus 4.8 by 3.66 and 3.00 points; and \ding{183} \textbf{Executability}: \ourcu{} reaches 69.86\% on FinSearchComp T2, outperforming Agents-A1-35B and Nex-N2-mini by 22.83 and 12.78 points, while \ourag{} achieves 76.00\% and 60.49\% on FinanceAgentBench v1.1 and v2, respectively. These results establish a path toward trustworthy financial intelligence in which domain expertise, long-horizon execution, and auditable evidence are jointly engineered as a unified foundation for frontier agentic models.

}

\reportdata[\textcolor{cyan}{\faGlobe}\enspace Website]{\url{https://mint-fin.github.io/mint-agent}}

\begin{document}
\maketitle

\begin{figure}[!ht]
    \centering
    \includegraphics[width=\linewidth]{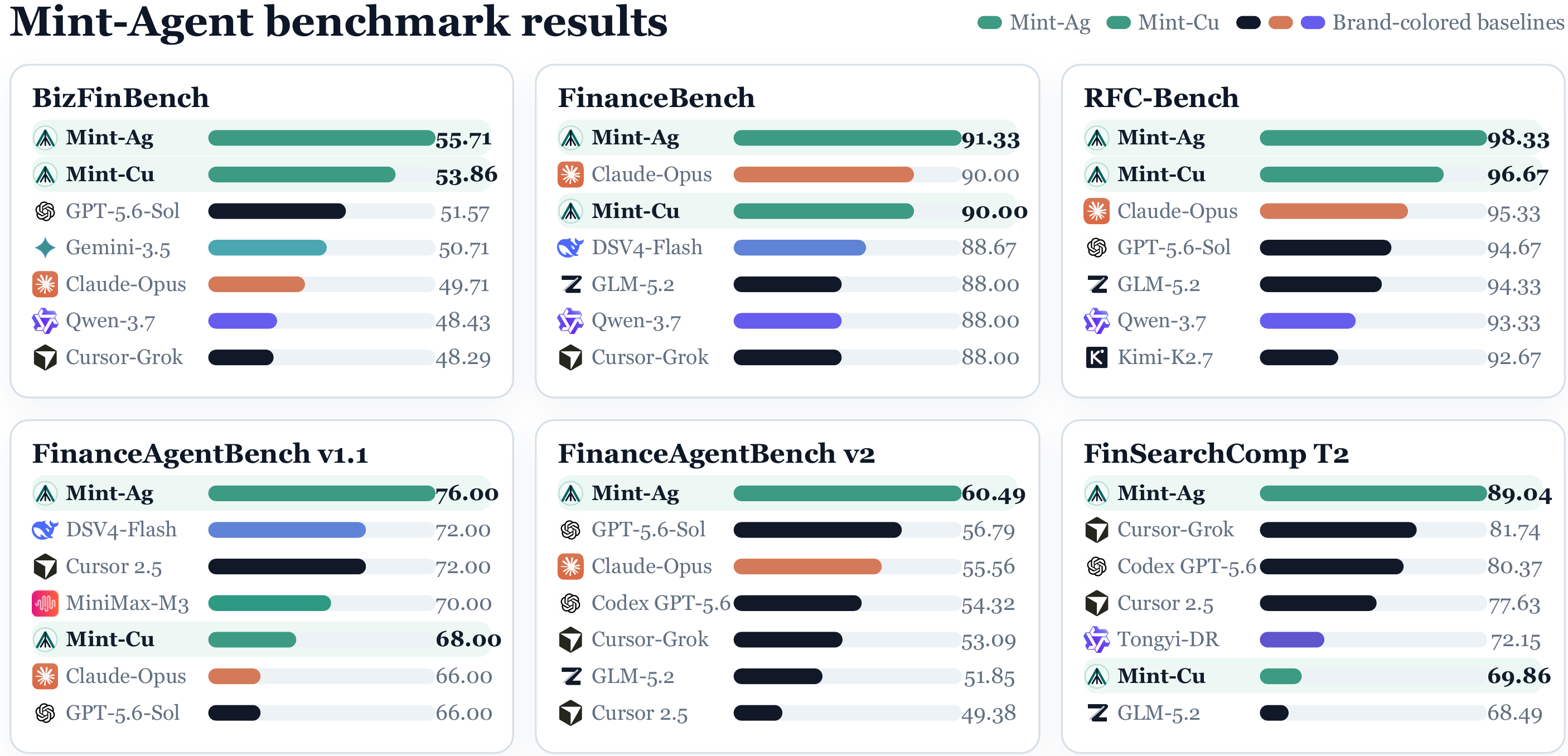}
    \vspace{-0.7em}
    \label{fig:benchmark-overview}
\end{figure}

\section{Introduction}
\label{sec:introduction}

Finance research is a demanding setting for autonomous agents because the answer is rarely a single fact. It often emerges from reading a disclosure in context, identifying the correct reporting period, comparing values across sources, and carrying the arithmetic into a decision. Existing agents can break this chain even when individual steps look plausible: a source may be relevant but not authoritative, a table may be read under the wrong fiscal period, or a computed figure may silently mix units. In finance, such errors are not merely inaccuracies; without a traceable route from source to calculation to conclusion, the result cannot be trusted or repaired.

This work argues that a capable financial agent should be optimized around one shared contract: every answer should be treated as a claim backed by recoverable evidence and reproducible calculation. If tasks are generated without provenance, the model may learn the surface form of financial analysis rather than the discipline behind it. If execution does not preserve state, the most informative failures disappear before they can become supervision. If training only rewards the final text, the decisions that determine whether a research trajectory succeeds remain under-specified. The problem is therefore to make the entire pipeline reinforce the same notion of financial correctness.

\begin{figure}
    \centering
    \includegraphics[width=0.8\linewidth]{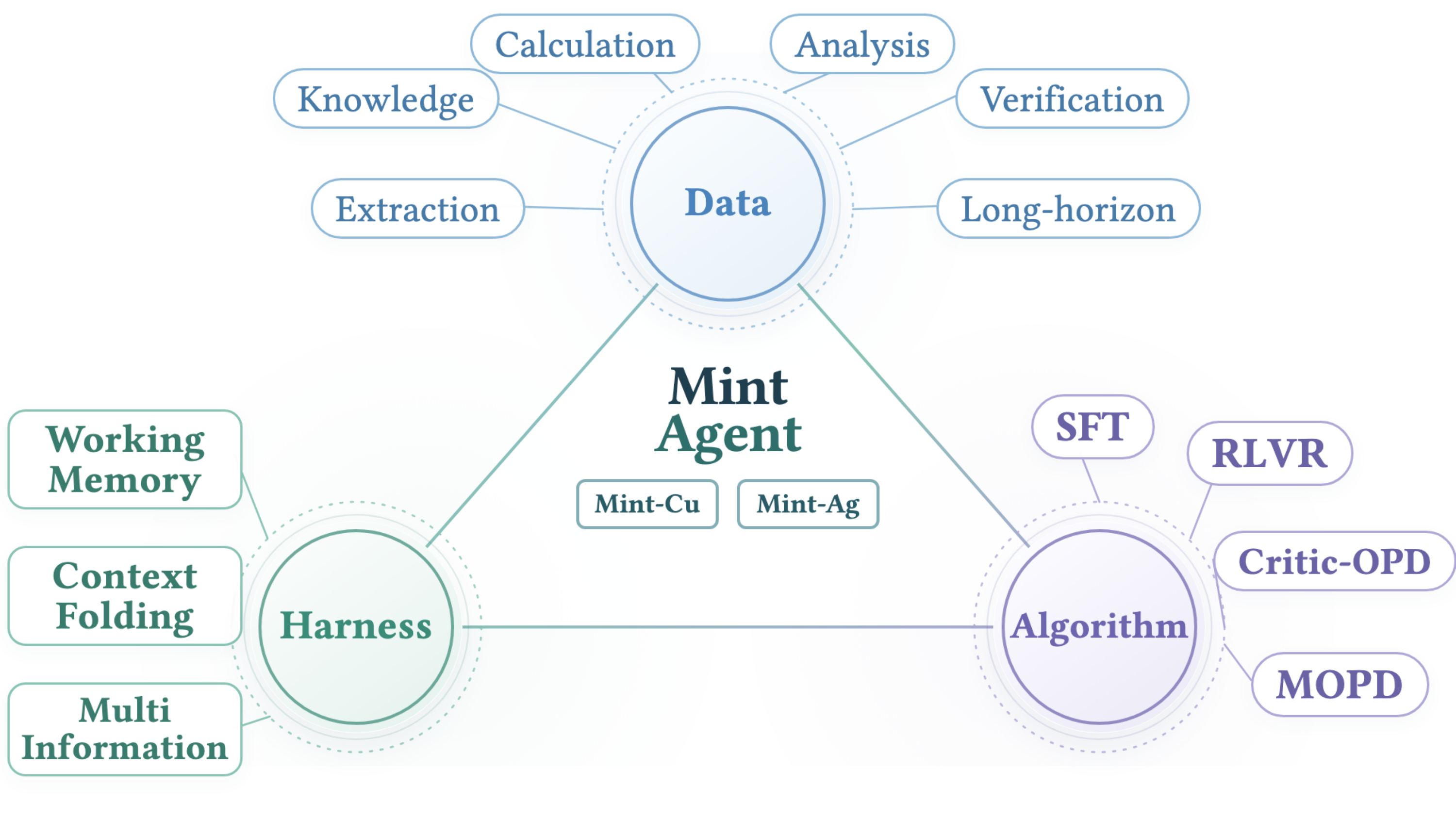}
    \caption{\textbf{Overview of the Mint Agent stack.}
Mint Agent is jointly developed along three complementary dimensions: data, harness, and algorithm. The data engine builds verifiable atomic and long-horizon financial tasks; the harness supports persistent context and multi-source interaction; and the algorithmic pipeline combines supervised fine-tuning, reinforcement learning with verifiable rewards, and on-policy distillation. Together, these components produce the Mint-Cu and Mint-Ag models.}
    \label{fig:intro2}
\end{figure}

To this end, we present \ourmethod, a series of finance-native agentic models for auditable financial research. \ourmethod develops capable and reliable financial intelligence through three complementary pillars spanning data, harness, and learning algorithms. \textbf{(I) Data}: Our data engine begins with real-world financial sources and constructs tasks that comprehensively evaluate financial data analysis, calculation, and reasoning. Through evidence crossover, it further synthesizes complex yet well-grounded multi-hop queries for financial research, testing financial tool use, long-horizon execution, and information seeking. \textbf{(II) Harness}: We introduce \textbf{MintHarness}, an evidence-first agent harness designed specifically for financial intelligence. It supports heterogeneous data sources and extended search trajectories while prioritizing evidence collection, thereby making the entire research process transparent and auditable. \textbf{(III) Algorithm}: Our training recipe first develops atomic financial capabilities and long-horizon agentic execution separately through reinforcement learning with verifiable rewards, and then integrates them through multi-teacher on-policy distillation. Together, these three pillars yield two unified financial agents, \textbf{\ourcu} and \textbf{\ourag}, with 9B and 27B parameters, respectively. Both models learn financial research as a recoverable sequential decision process rather than isolated answer generation. Despite its compact scale, \ourcu surpasses all counterparts in the $<$32B parameter class, while \ourag achieves performance competitive with substantially larger general-purpose models, including GPT-5.6-Sol and GLM-5.2.

We evaluate \ourmethod on seven professional financial benchmarks that jointly stress reliable financial reasoning and executable long-horizon research. Across this entire evaluation suite, \ourag{} consistently outperforms leading frontier models, including GPT-5.6-Sol, Claude Opus 4.8, and GLM-5.2, and attains the highest reported score on every benchmark. Specifically, it achieves 55.71\%, 91.33\%, and 98.33\% on BizFinBench, FinanceBench, and RFC-Bench, together with 76.00\% and 60.49\% on FinanceAgentBench v1.1 and v2 and 89.04\% and 54.07\% on FinSearchComp T2 and T3. Meanwhile, the compact \ourcu{} reaches 69.86\% on FinSearchComp T2, demonstrating that strong long-horizon financial execution can also be retained at the 9B scale. These results indicate that finance-native agentic models can combine domain reliability, extended execution, and auditability within one vertical intelligence stack.

Our main contributions are summarized as follows:

\begin{itemize}[leftmargin=*, noitemsep]
    \item \textbf{Full-stack finance-native intelligence.}
    We introduce a unified data--harness--algorithm framework that connects provenance-grounded task construction, persistent stateful execution, and verifiable policy learning under one recoverable evidence contract.
    \item \textbf{Flagship finance-native models.}
    We introduce \ourcu{} (9B) and \ourag{} (27B), two compact yet capable financial agents that combine reliable domain reasoning, long-horizon execution, and auditable evidence-grounded behavior.
    \item \textbf{Comprehensive empirical validation.}
    We evaluate on seven professional benchmarks and conduct cost, trajectory, failure-mode, and integration analyses, demonstrating leading financial reasoning, strong long-horizon execution, and auditable evidence-grounded behavior.
\end{itemize}

\section{Related Work}
\label{sec:related}

\subsection{Agentic Models}

Agentic models are emerging through a gradual transfer of control from hand-written workflows into the model policy: the harness still defines the environment and exposes actions, but post-training increasingly determines how an interaction unfolds. \textbf{Tool interaction.} What began as schema-constrained invocation is becoming a learned multi-turn policy that selects, sequences, and verifies actions against environmental feedback~\citep{agentlightning2025,agentr12025,muarl2025,rlfactory2025,trust2026,cm22026,deepagent2025,aworld2025}. \textbf{Planning and execution.} Credit assignment now reaches beyond isolated calls to the decisions that allocate effort, revise a plan, continue exploration, or terminate, allowing behavior to adapt to the state encountered along a trajectory~\citep{agentlightning2025,agentr12025,learningwhentoplan2025,aworld2025,internalizingfuture2026,selfevolvingwm2026,tongyidr2025,stepdr2025}. \textbf{Memory and context.} Long-horizon agency similarly shifts context management from passive retention to learned construction: policies decide what to preserve, recover, compress, and revisit as the raw interaction outgrows the context window~\citep{memoryr12025,memalpha2025,lookback2025,memsearcher2025,agenticmemory2026,qwenlongl152025,rlsummary2025,agentfold2025,resum2025}. \textbf{Recovery and coordination.} Reflection, progress estimation, and delegation extend the same principle to failure repair and multi-agent computation, making recovery paths and subagent allocation part of the policy rather than a fixed orchestration rule~\citep{retrospective2026,failurestronger2025,trust2026,reflectharness2026,yunque2026,dumate2026,selfoptimizingmas2026,multiorchestration2025,rao2026,unoorchestra2026}. As these abilities are trained separately, their integration becomes a further problem: weight-space methods cheaply compose related checkpoints but must resolve parameter interference, whereas on-policy distillation transfers specialist behavior on states generated by the student itself~\citep{yadav2023ties,gold2025,crossopd2026,mopd2026}. The emerging agent-native model is therefore not independent of its harness; it assumes more responsibility within an execution contract that must still preserve state, constrain actions, and expose failures for learning.

\subsection{Financial Intelligence}

Financial intelligence has progressed from knowing the language of finance toward performing work whose numerical and evidential boundaries are explicit. \textbf{From domain adaptation to grounded reasoning.} Financial pre-training and instruction tuning establish vocabulary, conventions, and broad task coverage, while more recent post-training targets quantitative reasoning and verifiable problem solving~\citep{bloomberggpt2023,fingpt2023,pixiu2023,finr12025}. The corresponding evaluation surface has expanded from calculations over supplied tables and text to disclosure-grounded question answering and the rejection of plausible but unsupported claims~\citep{finqa2021,tatqa2021,convfinqa2022,financebench2023,rfcbench2026}. This progression changes the standard of correctness: selecting the right entity and reporting period, preserving scale and unit, and recovering a valid derivation matter as much as matching the final value. \textbf{From answers to financial work.} A second transition removes the assumption that relevant evidence is already present. Analyst-style tasks require models to discover point-in-time information, reconcile professional documents, and carry quantitative conclusions across multiple sources, while execution-oriented settings test whether financial tools can be selected and applied under domain constraints~\citep{finsearchcomp2025,financeagentbench2025,findeepresearch2025,deepfinresearch2026,icbcbench2026,financecomplexqa2026,bigfinancebench2026,finresearchbenchii2026,fintoolbench2026,fintoolsyn2026,forcebench2026,cfagentbench2026,financeharness2026}. The unit of evaluation consequently moves from an isolated answer to a recoverable analyst workflow. This makes auditability a domain requirement rather than a presentation feature: a material conclusion is valid only when its source authority, time boundary, financial semantics, and calculation can be reconstructed, a requirement that neither domain adaptation nor a generic tool scaffold guarantees on its own.

\subsection{Deep Research}

Deep research operationalizes this wider notion of intelligence as an extended process of identifying what remains unknown, acquiring evidence, and revising a conclusion until the research objective is resolved. Two complementary lines have developed. \textbf{Harness-based systems.} Explicit research architectures maintain plans and intermediate state outside the model, coordinate parallel exploration, and assemble evidence before synthesis; this makes the trajectory observable and controllable, although performance remains coupled to a hand-designed topology~\citep{webweaver2025,scaffoldagent2026,argus2026,yunque2026,minddr2026,dumate2026,selfoptimizingmas2026,agentdisco2026}. \textbf{Training-based systems.} A second line internalizes the research loop through synthesized interactions and agentic post-training, first learning sustained search from long trajectories and then assigning denser supervision to plans, evidence structures, reflection, and context management~\citep{tongyidr2025,stepdr2025,openresearcher2026,openseeker2026,s1deepresearch2026,searchr1plus2026,quest2026,decomposer2026,deeprubric2026,metaresearcher2026,selfevolvingdr2026,resum2025,rlsummary2025,agentfold2025}. These routes improve autonomy, but autonomy alone does not make research trustworthy: fluent reports may misattribute a source, cite evidence that does not entail the claim, or follow plausible misinformation into a false conclusion~\citep{deeptrace2025,auditabledeepresearch2026,citednotverified2026,errorlocalization2026,misknow2026,drnoise2026}. Reliable deep research therefore requires the two lines to meet—the policy must learn how to investigate, while the harness must retain enough evidence and computation state to audit what it concludes. \ourmethod instantiates this joint requirement in finance by grounding task construction, execution, verification, and specialist training in the same recoverable evidence contract.

\section{Data Engine}
\label{sec:data-engine}

The Mint data engine develops financial executability at two scales. \emph{Atomic finance} concerns a bounded operation whose evidence is already available; \emph{long-horizon execution} concerns the policy that must discover and compose such evidence through interaction. We synthesize the corresponding streams $\mathcal{D}_{\mathrm{atom}}$ and $\mathcal{D}_{\mathrm{long}}$ separately, and denote their union by $\mathcal{D}_{\textsc{Mint}}$. Atomic tasks teach the model to complete a well-scoped piece of financial work from bounded evidence, while long-horizon tasks teach it to find that evidence and organize the required operations across an extended trajectory.

\begin{figure}[!tbp]
    \centering
    \includegraphics[width=0.98\linewidth]{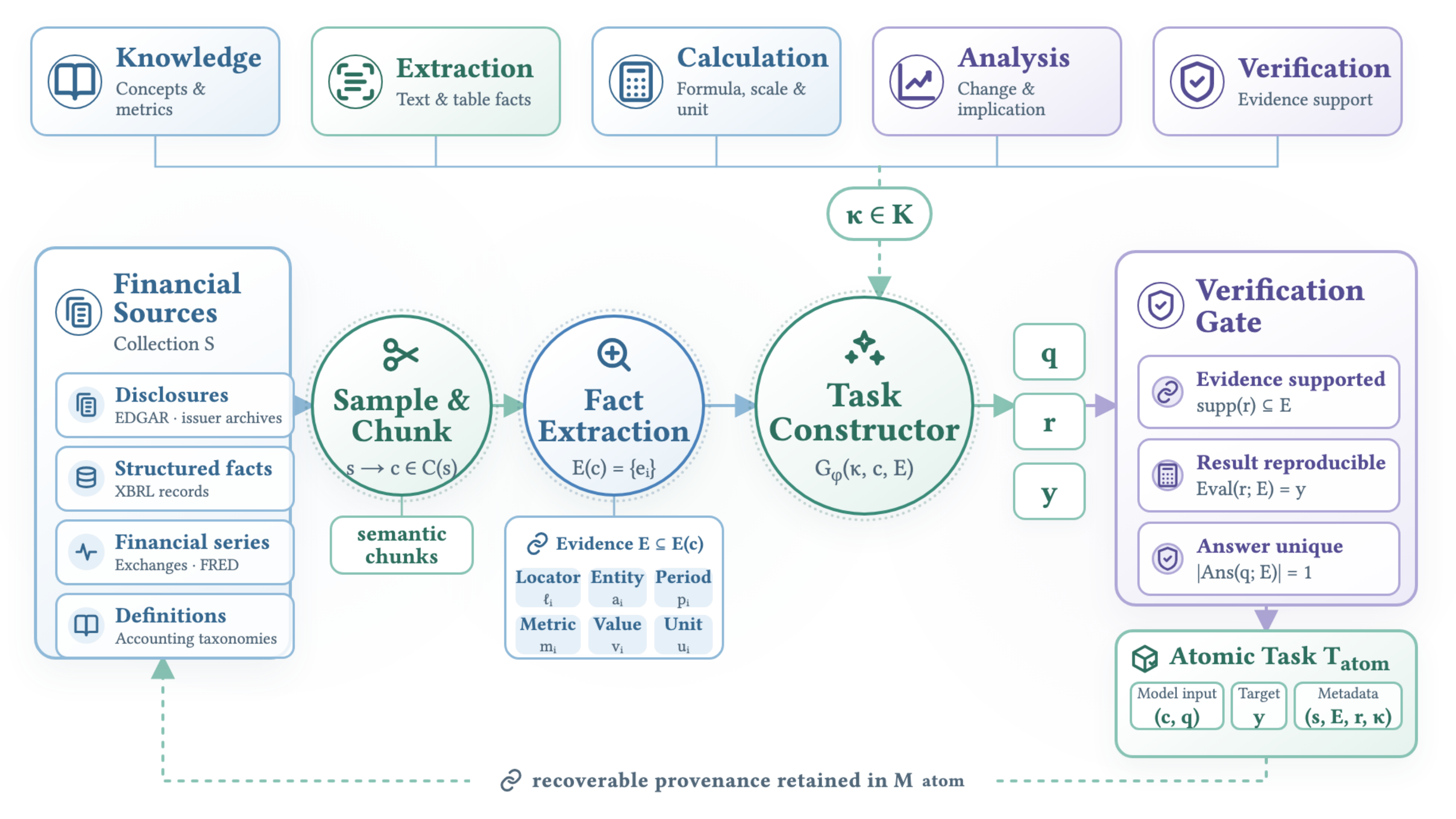}
    \caption{\textbf{Capability-guided construction of atomic financial tasks.}
A target capability is selected from knowledge, extraction, calculation, analysis, and verification. Primary financial sources are sampled and segmented into semantic chunks, from which structured, provenance-linked facts are extracted. A capability-conditioned constructor generates a question, reasoning trace, and answer, which are retained only after passing checks for evidential support, reproducibility, and answer uniqueness.}
    \label{fig:atomic-data-engine}
\end{figure}

\subsection{Atomic Financial Capabilities}
\label{sec:atomic-finance}

The first objective of the data engine is to build the model's atomic financial capabilities: the basic skills required to complete a self-contained financial task from a bounded context. We divide them into five groups:
\begin{itemize}[leftmargin=*, noitemsep]
    \item \textbf{Knowledge} covers financial concepts, conventions, and the meaning of commonly used metrics.
    \item \textbf{Extraction} identifies the fact requested by a question and recovers it from financial text or tables.
    \item \textbf{Calculation} applies a valid financial formula and returns a numerically correct result with the appropriate scale and unit.
    \item \textbf{Analysis} relates reported facts to explain a change, comparison, or financial implication.
    \item \textbf{Verification} determines whether a financial claim is supported by the given evidence.
\end{itemize}
Each task is assigned one primary capability $\kappa\in\mathcal{K}$, where $\mathcal{K}$ denotes the five groups above. This label controls what the question is intended to test; the financial topic of the question is recorded separately.

\paragraph{Data Sources.} We construct these tasks from primary financial materials rather than from existing question-answer datasets. Corporate disclosures are collected from EDGAR and issuer investor-relations archives, with their structured values aligned to the corresponding XBRL facts. Market observations come from exchange records, macroeconomic series from FRED, and financial definitions from the governing accounting taxonomies. Together, these sources form $\mathcal{S}$, while the taxonomies define the semantic environment $\Gamma_{\!\mathrm{fin}}$ used to check metric compatibility. Every source $s\in\mathcal{S}$ is stored with a stable locator and its reporting metadata, so any fact used during synthesis can be traced back to the exact document span or structured field from which it was obtained.

\begin{table}[t]
    \centering
    \small
    \caption{Sources used to construct atomic financial tasks.}
    \label{tab:atomic-data-sources}
    \begin{tabular}{p{0.24\linewidth}p{0.31\linewidth}p{0.35\linewidth}}
        \toprule
        \textbf{Source layer} & \textbf{Collection} & \textbf{Role} \\
        \midrule
        Disclosures & EDGAR and issuer archives & Grounded financial text and tables \\
        Structured facts & XBRL records & Normalized reported values \\
        Financial series & Exchange records and FRED & Market and macro observations \\
        Definitions & Accounting taxonomies & Metric semantics and valid formulas \\
        \bottomrule
    \end{tabular}
\end{table}

\paragraph{Task Construction.} Starting from $\mathcal{S}$, we sample a source $s$, divide it into semantically coherent chunks, and extract provenance-linked financial facts. Each fact records a source locator together with its entity, reporting period, metric, value, and unit; any fact whose supporting span cannot be recovered is removed. We then sample a target capability $\kappa$, select an evidence subset $E$ from a chunk $c$, and ask a capability-conditioned constructor $G_\phi$ to generate a question $q$, solution trace $r$, and answer $y$. A candidate is retained only if its derivation is fully supported, replayable, and unambiguous:
\[
    \operatorname{supp}(r)\subseteq E,
    \qquad
    \operatorname{Eval}(r;E)=y,
    \qquad
    \bigl|\operatorname{Ans}(q;E)\bigr|=1.
    \tag{1}
\]
where $\operatorname{supp}(r)$ is the evidence used by the trace, $\operatorname{Eval}(r;E)$ replays its financial operations, and $\operatorname{Ans}(q;E)$ is the set of answers admitted by the evidence. These checks anchor generation to the sampled source rather than to the constructor's parametric knowledge. The retained task is $\mathcal{T}_{\mathrm{atom}}=((c,q),y,\mathcal{M}_{\mathrm{atom}})$, with $\mathcal{M}_{\mathrm{atom}}=(s,E,r,\kappa)$ preserving the provenance and solution structure used for verification.

\begin{figure}
    \centering
    \includegraphics[width=0.95\linewidth]{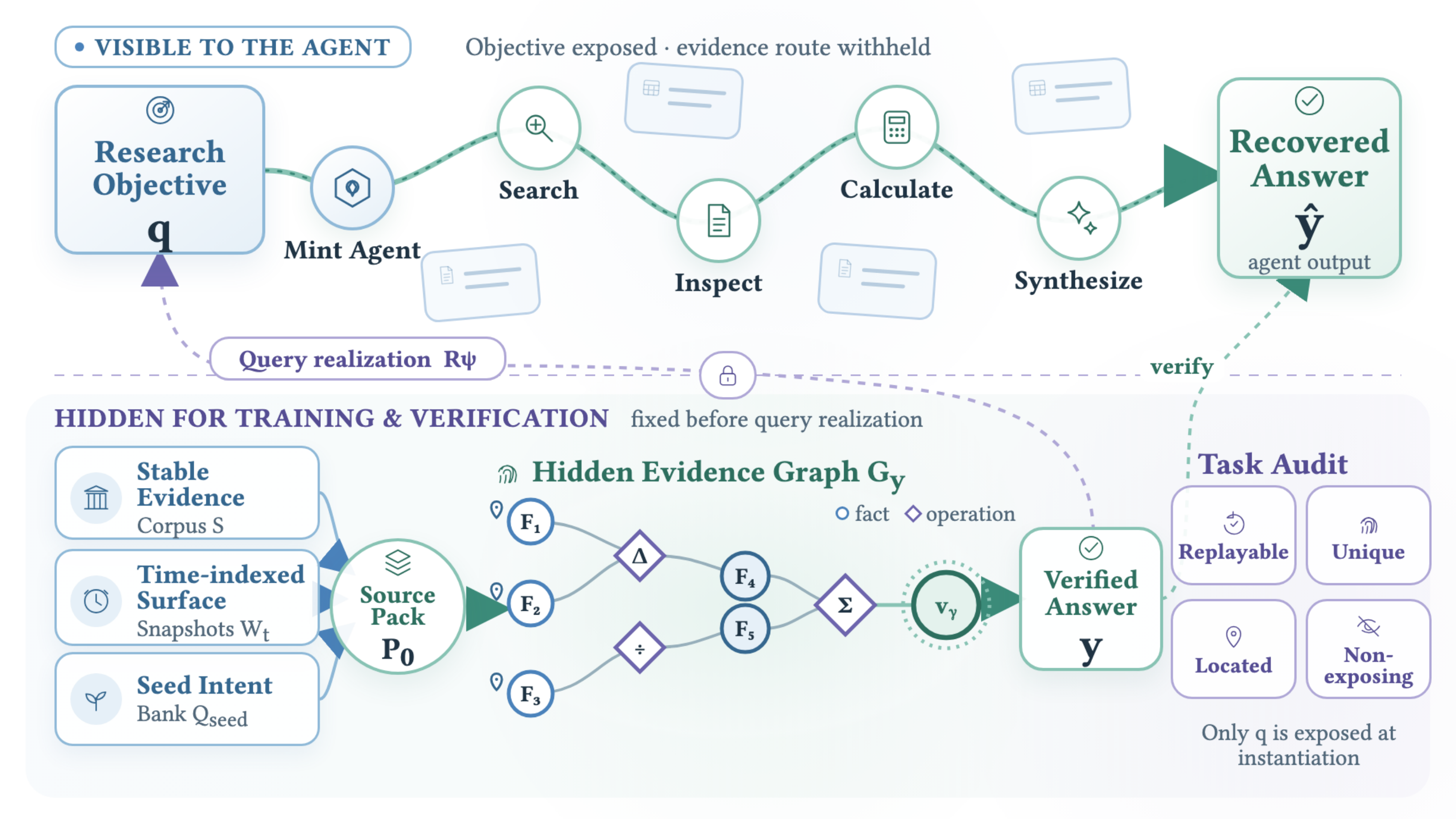}
    \caption{\textbf{Construction and verification of long-horizon agentic tasks.}
The agent observes only a research objective and must recover the answer through iterative search, inspection, calculation, and synthesis. The underlying source pack and evidence graph remain hidden, preserving the task's exploratory nature while enabling deterministic verification. }
    \label{fig:data-long}
\end{figure}

\subsection{Long-Horizon Agentic Execution}
\label{sec:long-horizon-execution}

After building the atomic capabilities, we turn to long-horizon execution. A long-horizon task provides a well-defined financial objective but withholds the supporting context, so that the model must learn to locate the relevant information and assemble it across an open-ended environment. When the task is instantiated in MintHarness, this information is recovered through an extended interaction with the real-world environment. The data engine therefore constructs each query together with a complete hidden answer graph, allowing that interaction to be trained and evaluated against a result fixed in advance.

\paragraph{Data Sources.} Long-horizon synthesis reuses the grounded corpus $\mathcal{S}$ from the atomic pipeline and extends it with a time-indexed financial surface $\mathcal{W}_{t_{\mathrm{cut}}}$, where $t_{\mathrm{cut}}$ denotes the collection cutoff. Each item in $\mathcal{W}_{t_{\mathrm{cut}}}$ is stored as a snapshot, so evidence drawn from a changing page or feed remains reproducible after its live version is updated. We also maintain a seed bank $\mathcal{Q}_{\mathrm{seed}}$ built from analyst-written requests after removing all overlap with the evaluation sets. 

\paragraph{Evidence Graph Construction.} We sample a seed query $q_0\in\mathcal{Q}_{\mathrm{seed}}$ and parse it into a task specification $z_0$ that preserves the required reasoning structure while discarding the original entities and values. The engine then mines $\mathcal{S}\cup\mathcal{W}_{t_{\mathrm{cut}}}$ for a new source pack $\mathcal{P}_0$, extracts its grounded facts into $E_0$, and composes a directed acyclic evidence graph $\mathbb{G}_0=(V_F\,\dot\cup\,V_O,A,\rho)$. Fact nodes $V_F$ contain extracted or derived values, operation nodes $V_O$ encode valid transformations under $\Gamma_{\!\mathrm{fin}}$, and the provenance map $\rho$ links every source-backed leaf to its exact locator in $\mathcal{P}_0$. The engine selects a terminal fact $v_y$ and retains its complete ancestor graph only when it meets minimum reasoning-depth and source-breadth requirements:
\[
    \mathbb{G}_y=\operatorname{Anc}(v_y;\mathbb{G}_0),
    \qquad
    y=\operatorname{Val}(v_y),
    \qquad
    \operatorname{depth}(\mathbb{G}_y)\geq h_{\min},
    \qquad
    \bigl|\operatorname{Src}(\mathbb{G}_y)\bigr|\geq b_{\min},
    \tag{2}
\]
where $\operatorname{Anc}(v_y;\mathbb{G}_0)$ is the subgraph containing $v_y$ and all nodes required to derive it, $\operatorname{Val}(v_y)$ is the value carried by the terminal fact, $\operatorname{depth}(\mathbb{G}_y)$ counts the greatest number of operation nodes on a leaf-to-terminal path, and $\operatorname{Src}(\mathbb{G}_y)$ is the set of source records supporting its leaves. The thresholds $h_{\min}$ and $b_{\min}$ impose the minimum reasoning depth and source breadth, respectively. Every operation in $\mathbb{G}_y$ is then replayed. This construction fixes a nontrivial answer and its complete derivation before a new query is written.

\paragraph{Query Realization and Audit.} A query constructor $R_\psi$ realizes $q$ from the seed intent $q_0$ and the sampled pair $(\mathbb{G}_y,y)$: the seed determines the task form, while the graph determines its new financial content. A candidate passes the audit only when replay returns the fixed answer, the answer is unique, every leaf has localized provenance, and the query does not expose its hidden route:
\[
    \operatorname{Eval}(\mathbb{G}_y)=y,
    \qquad
    \operatorname{Ans}(q;\mathbb{G}_y)=\{y\},
    \qquad
    \operatorname{Loc}(\mathbb{G}_y)=1,
    \qquad
    \operatorname{Expose}(q;\mathbb{G}_y)=0.
    \tag{3}
\]
where $\operatorname{Loc}$ checks the provenance map and $\operatorname{Expose}$ detects leaked locators or intermediate facts. We additionally reject any query that reproduces its seed or overlaps an evaluation task, and require time-sensitive queries to state the boundary $t_{\mathrm{cut}}$ under which their answer is valid. The retained task is $\mathcal{T}_{\mathrm{long}}=(q,y,\mathcal{M}_{\mathrm{long}})$, where $\mathcal{M}_{\mathrm{long}}=(q_0,t_{\mathrm{cut}},\mathcal{P}_0,\mathbb{G}_y)$. Only the self-contained objective $q$ is exposed during execution; the answer and audit metadata remain hidden.

\begin{figure}[!t]
    \centering
    \includegraphics[width=\linewidth]{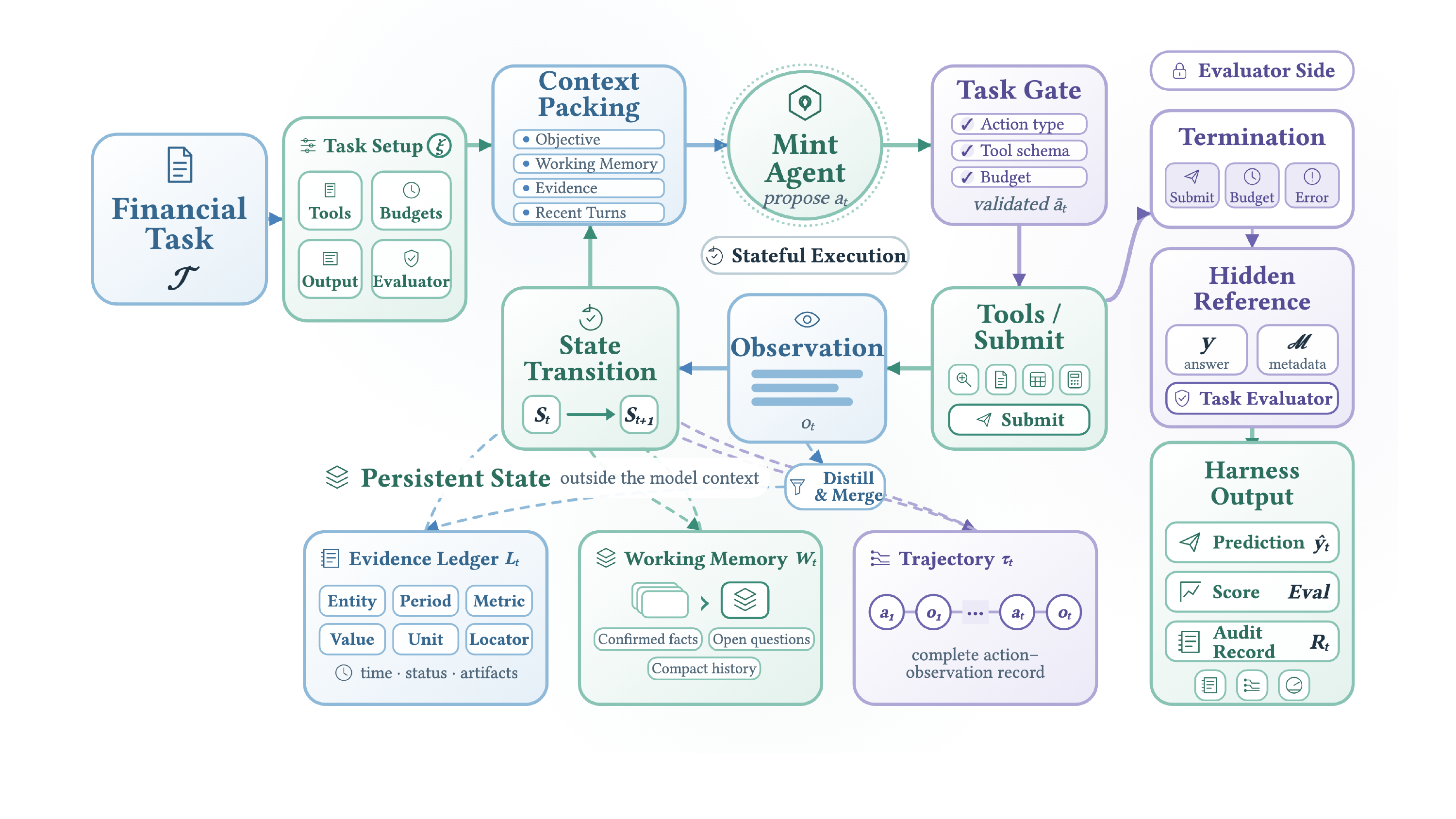}
    \vspace{-4em}
    \caption{\textbf{MintHarness overview.} Given a financial task and task-specific configuration, MintHarness orchestrates a stateful execution loop in which the agent proposes actions, the harness validates and executes tool calls, and resulting observations update the execution state. Evidence, working memory, and the complete action–observation trajectory are maintained outside the bounded model context. Upon submission, budget exhaustion, or error, an isolated evaluator compares the prediction against hidden reference data and produces a score and auditable execution record.}
    \label{fig:harness}
\end{figure}

\section{MintHarness}
\label{sec:mint-harness}

MintHarness instantiates bounded financial questions and open-ended research objectives as a common stateful execution process. It normalizes each task, exposes a task-specific action space, records the interaction, and maintains persistent state outside the model context.

\subsection{Execution Loop}
\label{sec:harness-loop}

\paragraph{Task Setup.} Let $\xi$ denote the configuration of a task $\mathcal{T}$, specifying its tools, budgets, output format, and evaluator. MintHarness maps $\mathcal{T}$ to a visible input $x_{\mathcal{T}}$ and an admissible action set $\mathcal{A}_{\xi}$. For an atomic task, $x_{\mathcal{T}}=(c,q)$ and $\mathcal{A}_{\xi}=\{\textsc{Submit}\}$; for a long-horizon task, $x_{\mathcal{T}}=q$ and $\mathcal{A}_{\xi}=\mathcal{U}_{\xi}\cup\{\textsc{Submit}\}$, where $\mathcal{U}_{\xi}$ contains the configured tools. These objects initialize the harness state:
\begin{equation}
    \mathbf{S}_0=\operatorname{Init}_{\xi}(\mathcal{T})
    \equiv\bigl(x_{\mathcal{T}},\mathcal{A}_{\xi},\mathbf{b}_0,
    \mathbf{L}_0,\mathbf{W}_0,\boldsymbol{\tau}_0\bigr),
    \tag{4}
\end{equation}
where $\mathbf{b}_0$ contains the initial step, tool, and context budgets, while $\mathbf{L}_0$, $\mathbf{W}_0$, and $\boldsymbol{\tau}_0$ initialize the evidence ledger, working memory, and trajectory as empty objects. The reference answer $y$ and verification metadata $\mathcal{M}$ remain on the evaluator side and never enter $\mathbf{S}_0$.

\paragraph{Action Cycle.} At turn $t$, $\operatorname{Pack}_{B}(\mathbf{S}_t)$ constructs the model-visible context under token budget $B$ from the task, compact working memory, relevant ledger entries, and recent turns. The policy proposes an action from this view, and the harness maps it to either an executable action or an explicit error through the task gate:
\begin{equation}
    a_t\sim\pi_{\theta}\!\left(\,\cdot\mid
    \operatorname{Pack}_{B}(\mathbf{S}_t)\right),
    \qquad \bar a_t=\operatorname{Gate}_{\xi}(a_t;\mathbf{S}_t),
    \tag{5}
\end{equation}
where $\operatorname{Gate}_{\xi}$ checks the action type, tool schema, and remaining execution budget. This validation changes only whether a proposed action can be executed; the research decision itself remains with the policy.

\paragraph{Evidence Update.} If $\bar a_t\in\mathcal{U}_{\xi}$, let $r_t=\mathcal{U}_{\bar a_t}(\operatorname{args}(\bar a_t))$ be the raw tool result and set $o_t=r_t$. MintHarness distills this result into localized financial records before merging them into the ledger:
\begin{equation}
    \mathbf{L}_{t+1}
    =\operatorname{Merge}\!\left(
    \mathbf{L}_t,\operatorname{Distill}(r_t;x_{\mathcal{T}});\nu_t\right),
    \qquad \bar a_t\in\mathcal{U}_{\xi},
    \tag{6}
\end{equation}
where $\operatorname{Distill}$ produces typed, provenance-linked financial records, and $\nu_t$ retains acquisition time, tool status, and artifact references. A failed call contributes no evidence, but its error-valued observation remains available to the next turn. Submission and error actions likewise produce an observation $o_t$ while leaving $\mathbf{L}_{t+1}=\mathbf{L}_t$.

\paragraph{State Transition.} The transition operator appends $(\bar a_t,o_t)$ to the trajectory, advances the remaining budgets, and refreshes the working memory from the updated trajectory and ledger. The ledger remains persistent even when only a selected view is packed into the next prompt:
\begin{equation}
    \mathbf{S}_{t+1}
    =F_{\xi}(\mathbf{S}_t;\bar a_t,o_t,\mathbf{L}_{t+1}),
    \tag{7}
\end{equation}
where $F_{\xi}$ is the configured state transition. Equations~(5)--(7) repeat until the model submits an answer, the step budget is exhausted, or an unrecoverable environment error occurs.

\paragraph{Completion.} At termination, $\hat y_T$ is the submitted answer, or $\bot$ if execution ends without one. Let $\mathbf{R}_T=(\mathbf{L}_T,\boldsymbol{\tau}_T,\operatorname{Usage}_T,\operatorname{Status}_T)$ collect the auditable execution record. The task-specific evaluator receives only the extracted prediction and isolated reference answer:
\begin{equation}
    \operatorname{Out}_{\xi}(\mathcal{T})
    =\bigl(\hat y_T,\operatorname{Eval}_{\xi}(\hat y_T;y),\mathbf{R}_T\bigr).
    \tag{8}
\end{equation}
The output includes the prediction, score, and execution record.

\subsection{Module Designs}
\label{sec:harness-design}

\paragraph{Evidence Ledger.} Equation~(6) stores distilled facts as persistent typed records with source locators, financial scope, and acquisition metadata. The ledger also records failed retrieval paths separately from supporting evidence, preserving the distinction between missing support and inaccessible sources.

\paragraph{Working Memory.} The working memory $\mathbf{W}_t$ is a persistent but compact semantic state that carries the investigation across turns. After each state transition, it is refreshed from the trajectory and evidence ledger to retain the current plan, confirmed conclusions, unresolved questions, and intermediate calculations, with factual claims linked back to their ledger records. 

\paragraph{Context Management.} Equations~(5) and~(7) separate durable execution state from the bounded context shown to the model. Recent turns remain verbatim, older interactions are compressed into working memory, and long documents or tables remain accessible through artifact references. The packing function selects the objective, relevant state, and ledger entries for each turn under the token budget.

\section{Training Pipeline}
\label{sec:training-pipeline}

\begin{figure}[!tbp]
    \centering
    \includegraphics[width=0.95\linewidth]{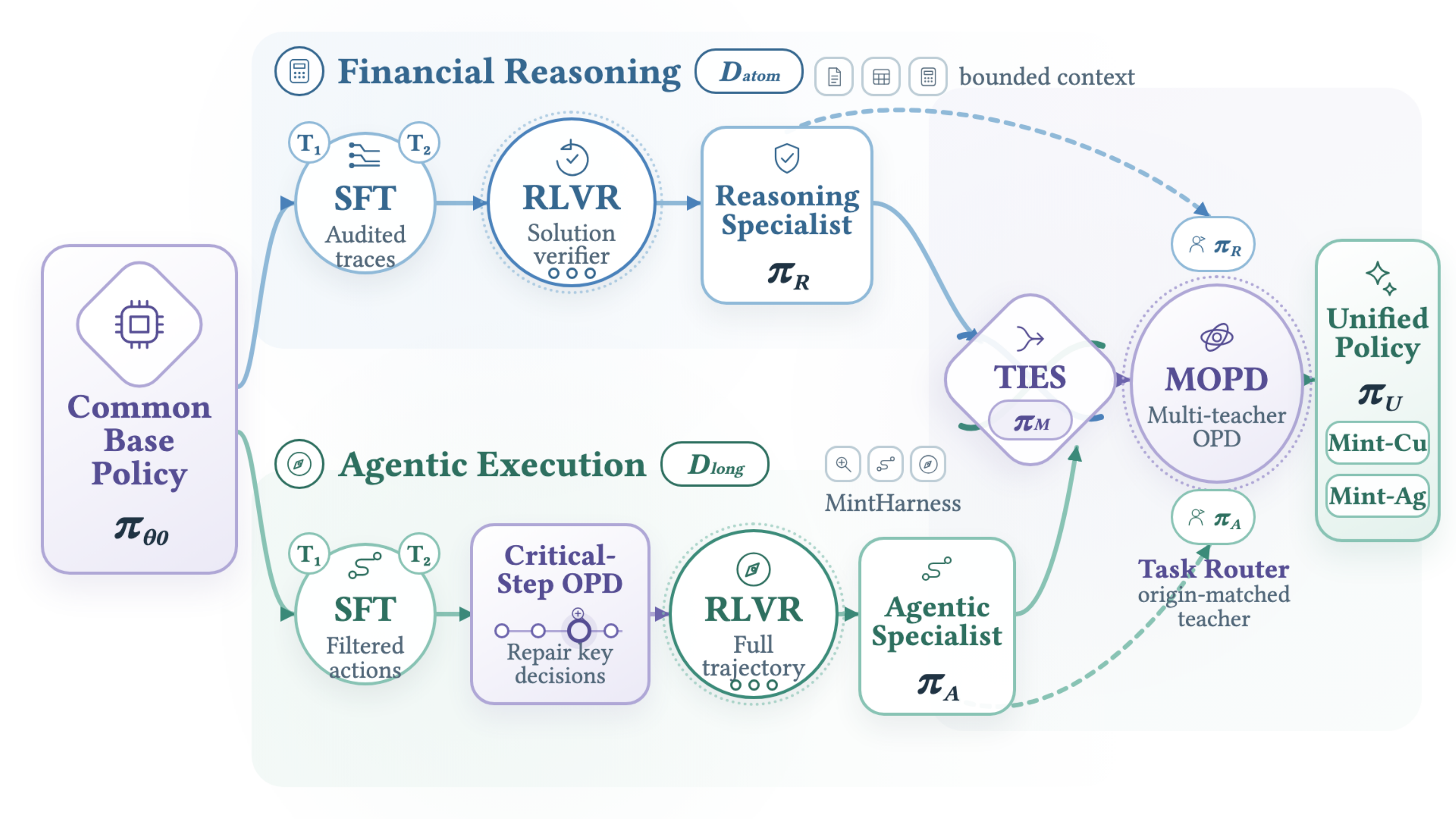}
    \caption{\textbf{Twin-specialist training and integration pipeline.}
Starting from a shared base policy, we train a financial-reasoning specialist on atomic tasks using supervised fine-tuning and reinforcement learning with solution-level verifiers, and an agentic-execution specialist on long-horizon tasks using supervised fine-tuning, critical-step on-policy distillation, and trajectory-level reinforcement learning. The two specialists are first merged with TIES and then integrated through multi-teacher on-policy distillation with origin-matched routing, producing a unified policy instantiated as Mint-Cu and Mint-Ag.}
    \label{fig:training-pipeline}
\end{figure}

Starting from a common base policy $\pi_{\theta_0}$, we train a financial-reasoning specialist on $\mathcal{D}_{\mathrm{atom}}$ and an agentic-execution specialist on $\mathcal{D}_{\mathrm{long}}$, then integrate them into a unified policy.

\subsection{Financial Reasoning}
\label{sec:financial-reasoning-training}

The financial reasoning branch teaches the model to execute the five atomic capabilities defined in \Cref{sec:atomic-finance} when the relevant context is already available. Its input remains $(c,q)$, and its response contains a financial derivation followed by the answer. We train this branch in two stages: SFT first anchors the policy to valid solution forms, after which RLVR explores alternative derivations and optimizes them directly against the hidden evidence and solution structure carried by $\mathcal{M}_{\mathrm{atom}}$.

\paragraph{SFT.} For every $\mathcal{T}_{\mathrm{atom}}\in\mathcal{D}_{\mathrm{atom}}$, we prompt GLM-5.2~\citep{glm522026} and DeepSeek-V4-Pro~\citep{deepseekv42026} with $(c,q)$ and sample multiple candidate traces, each containing a complete financial derivation and its final answer. We pass these traces through three gates before using them as supervision. \textbf{Correctness.} The answer must agree with $y$, and the derivation must be replayable against $E$ under $\Gamma_{\!\mathrm{fin}}$. \textbf{Pattern filtering.} Deterministic rules reject a trace when it repeats the same step, breaks before a usable answer, or spends most of its length restating the prompt. \textbf{Trace selection.} If several candidates remain for the same input, a separate LLM judge selects the strongest one under a fixed rubric. The rubric favors a complete financial derivation, penalizes unsupported jumps, and prefers the clearest trace among equally correct alternatives. The selected teacher traces, together with the audited trace $r$ stored by the data engine, form the response targets for one round of supervised fine-tuning from $\pi_{\theta_0}$. We denote the resulting checkpoint by $\pi_{\mathrm{R}}^{\mathrm{SFT}}$, where the subscript $\mathrm{R}$ identifies the financial reasoning branch.

\paragraph{Verifier.} With the supervised checkpoint in place, we turn to RLVR. Its first requirement is a verifier that converts the audit information retained by the data engine into a training reward. Let $o=(\hat r,\hat y)$ denote a rollout containing a generated derivation $\hat r$ and final answer $\hat y$. For $\mathcal{T}_{\mathrm{atom}}=((c,q),y,\mathcal{M}_{\mathrm{atom}})$ with $\mathcal{M}_{\mathrm{atom}}=(s,E,r,\kappa)$, we define
\[
    R_{\mathrm{atom}}(o,\mathcal{T}_{\mathrm{atom}})
    =\mathsf{A}_{\kappa}(\hat y,y)\,
     \mathsf{S}(\hat r;s,E)\,
     \mathsf{C}_{\kappa}(\hat r,\hat y;r,\Gamma_{\!\mathrm{fin}})
    \in\{0,1\},
    \tag{9}
\]
where $\mathsf{A}_{\kappa}$ is the capability-aware answer-equivalence check, $\mathsf{S}$ verifies that every material premise in $\hat r$ is supported by a locator in $E$ and its source $s$, and $\mathsf{C}_{\kappa}$ replays the derivation under $\Gamma_{\!\mathrm{fin}}$, including its period and unit constraints. The reference trace $r$ specifies the required financial relation but does not impose a particular wording or path, so an alternative derivation receives credit whenever it is grounded, internally consistent, and reaches the verified answer.

\paragraph{RLVR.} Given this verifier, we initialize RLVR from $\pi_{\mathrm{R}}^{\mathrm{SFT}}$ and organize the training set as a difficulty-aware curriculum. For a task $\mathcal{T}_j\in\mathcal{D}_{\mathrm{atom}}$ with input $(c_j,q_j)$, we draw ten independent rollouts from the SFT checkpoint and let $p_j^{(0)}$ be the fraction accepted by $R_{\mathrm{atom}}$. We retain the initial RL pool $\mathcal{D}_{\mathrm{R}}^{(0)}=\{\mathcal{T}_j:p_j^{(0)}\leq0.8\}$; tasks solved in more than eight trials are already stable and do not enter RL. Difficulty is then refreshed online rather than frozen at this initial estimate. At update $m$, the old policy $\pi_{\theta_{m-1}}$ produces a group of $G$ responses for each candidate task, giving
\[
    p_j^{(m)}=\frac{1}{G}\sum_{g=1}^{G}
    R_{\mathrm{atom}}\!\left(o_{j,g},\mathcal{T}_j\right),
    \qquad
    \mathcal{B}_m
    =\left\{\mathcal{T}_j\in\mathcal{D}_{\mathrm{R}}^{(0)}:
    0<p_j^{(m)}<1\right\}.
    \tag{10}
\]
Here, $o_{j,g}\sim\pi_{\theta_{m-1}}(\cdot\mid c_j,q_j)$ is the $g$-th rollout, $p_j^{(m)}$ is the group's verified pass rate, and $\mathcal{B}_m$ is the active batch. Candidate tasks are oversampled until $\mathcal{B}_m$ is full; groups with all-zero or all-one rewards are deferred because they provide no relative learning signal. Within each active group, we normalize the verified rewards and optimize them with sequence-level GSPO~\citep{gspo2025}, whose importance ratio is the geometric mean of token likelihood ratios under the current and rollout policies. This makes each update correspond to the complete financial solution scored by the verifier. The optimized checkpoint, denoted $\pi_{\mathrm{R}}$, has therefore learned the atomic capabilities from verified execution rather than from teacher traces alone.

\subsection{Financial Agentic Execution}
\label{sec:financial-agentic-training}

Financial agentic execution trains the model to keep working when the evidence required by a query is not given in advance. Within MintHarness, the agent must collect information over an extended interaction, separate useful evidence from noise, and carry the surviving evidence into an answer that can be reproduced and audited. We develop this policy in three stages. SFT first establishes stable interaction behavior; a targeted on-policy distillation stage then repairs consequential decisions exposed by that checkpoint; finally, RLVR optimizes complete trajectories against the hidden answer graph.

\paragraph{SFT.} For each $\mathcal{T}_{\mathrm{long}}=(q,y,\mathcal{M}_{\mathrm{long}})$ in $\mathcal{D}_{\mathrm{long}}$, we use GLM-5.2 and DeepSeek-V4-Pro to execute $q$ in MintHarness and collect trajectories $\tau=(a_1,o_1,\ldots,a_T,o_T,\hat y)$, where $a_t$ and $o_t$ are the action and environment observation at turn $t$, and $\hat y$ is the final answer. We filter this data at two levels. \textbf{Trajectory gate.} A trajectory is retained only when $\hat y$ agrees with $y$ and its evidence can be replayed against the source pack $\mathcal{P}_0$ and answer graph $\mathbb{G}_y$ stored in $\mathcal{M}_{\mathrm{long}}$. \textbf{Turn gate.} Within a retained trajectory, we remove a turn from the SFT loss if it repeats an action without acquiring new evidence, makes a call that contributes nothing to the answer graph, or invokes a tool that does not exist in MintHarness. The surrounding history and observations remain in the conditioning context, so filtering a turn does not break the state seen by later actions. We fine-tune $\pi_{\theta_0}$ only on the retained model actions and final answers; environment observations are never prediction targets. The resulting checkpoint is denoted $\pi_{\mathrm{A}}^{\mathrm{SFT}}$, where $\mathrm{A}$ identifies the agentic branch.

\paragraph{Critical-Step OPD.} SFT provides a viable interaction policy, but its failed trajectories reveal decisions for which sparse outcome supervision is too coarse. We roll out $\pi_{\mathrm{A}}^{\mathrm{SFT}}$ on $\mathcal{D}_{\mathrm{long}}$ and ask an LLM judge to identify critical turns $\mathcal{I}(\tau^{-})$ in each failed trajectory $\tau^{-}$. A turn is critical when changing its action can restore a valid route to $\mathbb{G}_y$ without altering the preceding interaction. For every selected turn, we preserve the pre-turn state $h_t=(q,a_1,o_1,\ldots,a_{t-1},o_{t-1})$; the current student resamples the next action from $h_t$, and a frozen DeepSeek-V4-Pro teacher scores that same state and student continuation. Distillation therefore concentrates on consequential decisions rather than imitating complete teacher trajectories~\citep{mopd2026}.

Because the teacher and student use different tokenizers, we align their tokenizations into minimal chunks with identical decoded text following GOLD-style cross-tokenizer OPD~\citep{gold2025,crossopd2026}. Let $z_{t,1:L_t}$ be the student action tokens. For a matched chunk $c$, $C^{\mathrm{S}}_{t,c}$ denotes its student-token positions, while $\ell^{\mathrm{S}}_{t,c}$ and $\ell^{\mathrm{T}}_{t,c}$ are the chunk log-likelihoods under the rollout student $\pi_{\theta_{\mathrm{old}}}$ and teacher $\pi_{\mathrm{T}}$. We distribute the teacher's chunk log-likelihood across the aligned student tokens and define the resulting token-level advantage as
\[
    \begin{aligned}
    \widetilde\ell^{\mathrm{T}}_{t,l}
    &=\frac{\ell^{\mathrm{T}}_{t,c}}{\ell^{\mathrm{S}}_{t,c}}
      \log\pi_{\theta_{\mathrm{old}}}(z_{t,l}\mid h_t,z_{t,<l}),\\
    \widehat A^{\mathrm{OPD}}_{t,l}
    &=\operatorname{sg}\!\left[
      \widetilde\ell^{\mathrm{T}}_{t,l}
      -\log\pi_{\theta_{\mathrm{old}}}(z_{t,l}\mid h_t,z_{t,<l})
      \right],
      \qquad l\in C^{\mathrm{S}}_{t,c},
    \end{aligned}
    \tag{11}
\]
where $\widetilde\ell^{\mathrm{T}}_{t,l}$ is the teacher target assigned to student token $l$ and $\operatorname{sg}$ stops gradients through the advantage. We optimize these advantages over the selected action tokens with the standard clipped token-level OPD surrogate, using the current-to-rollout student likelihood ratio. This yields $\pi_{\mathrm{A}}^{\mathrm{OPD}}$, correcting the student precisely where its own execution first becomes consequentially wrong.

\paragraph{RLVR.} We then optimize $\pi_{\mathrm{A}}^{\mathrm{OPD}}$ through full interactions with MintHarness. For a rollout $\tau$ of $\mathcal{T}_{\mathrm{long}}=(q,y,\mathcal{M}_{\mathrm{long}})$, where $\mathcal{M}_{\mathrm{long}}=(q_0,t_{\mathrm{cut}},\mathcal{P}_0,\mathbb{G}_y)$, the trajectory reward is
\[
    R_{\mathrm{long}}(\tau,\mathcal{T}_{\mathrm{long}})
    =\mathsf{A}_{\mathrm{long}}(\hat y,y)\,
     \mathsf{G}(\tau;\mathcal{P}_0,\mathbb{G}_y)\,
     \mathsf{H}(\tau;t_{\mathrm{cut}})
    \in\{0,1\},
    \tag{12}
\]
where $\mathsf{A}_{\mathrm{long}}$ checks the final answer, $\mathsf{G}$ verifies that the evidence recovered in $\tau$ can replay the required path through $\mathbb{G}_y$ with provenance in $\mathcal{P}_0$, and $\mathsf{H}$ checks that the interaction uses valid MintHarness actions and respects the collection cutoff $t_{\mathrm{cut}}$. Thus, a correct-looking answer receives no reward when its supporting route cannot be recovered. For every query, the policy samples a group of $G$ complete trajectories. We reuse the active-batch rule in Eq.~(10), replacing $R_{\mathrm{atom}}$ with $R_{\mathrm{long}}$, and apply the same sequence-level GSPO update described above. The likelihood ratio covers model-generated reasoning, actions, and the final answer across the complete interaction; environment observations remain conditioning state and are excluded from the loss. The resulting specialist $\pi_{\mathrm{A}}$ learns to sustain financial research until it produces an answer whose evidence and execution can both be audited.

\subsection{Expert Integration}
\label{sec:expert-integration}

We integrate the reasoning expert $\pi_{\mathrm{R}}$ and execution expert $\pi_{\mathrm{A}}$ through TIES merging followed by multi-teacher on-policy distillation.

\paragraph{TIES Model Merging.} Both experts descend from the common base policy $\pi_{\theta_0}$. We apply TIES~\citep{yadav2023ties} to their task vectors, defined as the parameter displacements of $\pi_{\mathrm{R}}$ and $\pi_{\mathrm{A}}$ from that base. TIES first trims low-magnitude coordinates from each task vector, then elects a shared sign at every coordinate and discards expert updates that disagree with it. The remaining aligned updates are averaged, scaled, and added back to the base parameters; coordinates with no surviving update retain their base value. The resulting initialization $\pi_{\mathrm{M}}$ therefore preserves the experts' dominant compatible directions while suppressing direct sign conflicts. Weight-space compatibility alone, however, does not guarantee that each expert's behavior is retained on the states where it matters.

\paragraph{Multi-Teacher On-Policy Distillation.} We therefore initialize a student $\pi_{\theta}$ from $\pi_{\mathrm{M}}$ and retain $\pi_{\mathrm{R}}$ and $\pi_{\mathrm{A}}$ as frozen teachers. MOPD~\citep{mopd2026} assigns supervision by task origin rather than asking the two teachers to compete on every sample. Specifically, we define the training mixture and its teacher router as
\begin{equation}
    \mathcal{D}_{\mathrm{mix}}
    =\alpha_{\mathrm{R}}\mathcal{D}_{\mathrm{atom}}
    +\alpha_{\mathrm{A}}\mathcal{D}_{\mathrm{long}},
    \qquad
    \pi^{\mathrm{T}}_{\mathcal{T}}=
    \begin{cases}
        \pi_{\mathrm{R}}, & \mathcal{T}\in\mathcal{D}_{\mathrm{atom}},\\
        \pi_{\mathrm{A}}, & \mathcal{T}\in\mathcal{D}_{\mathrm{long}},
    \end{cases}
    \qquad
    \alpha_{\mathrm{R}}+\alpha_{\mathrm{A}}=1,
    \tag{13}
\end{equation}
where $\alpha_{\mathrm{R}}$ and $\alpha_{\mathrm{A}}$ are the sampling probabilities of the two empirical task distributions, $\mathcal{T}$ is a sampled task, and $\pi^{\mathrm{T}}_{\mathcal{T}}$ is its routed teacher. Thus, atomic problems receive the financial reasoning signal, whereas long-horizon interactions receive the agentic execution signal.
For a sampled task $\mathcal{T}$, the student generates model tokens $z_{1:L}$ on-policy, where $L$ is the rollout length. Let $h_t$ denote the complete conditioning state before token $z_t$: it contains the bounded financial input for an atomic task, or the accumulated interaction history and environment observations for a long-horizon task. The routed teacher is prefixed on these same student-generated states, and MOPD minimizes the token-averaged reverse KL from the student to that teacher. Since both specialists share the same base architecture and tokenizer, their distributions align token by token. We optimize the policy-gradient form of this objective; if $\pi_{\theta^-}$ is the student snapshot that produced the rollout, the teacher--student log-probability difference supplies a dense advantage at every generated token:
\begin{equation}
    \begin{aligned}
        A^{\mathrm{MOPD}}_t
        &=\operatorname{sg}\!\left[
        \log\pi^{\mathrm{T}}_{\mathcal{T}}(z_t\mid h_t)
        -\log\pi_{\theta^-}(z_t\mid h_t)
        \right],\\
        \mathcal{L}_{\mathrm{MOPD}}(\theta)
        &=-\mathbb{E}\!\left[
        \frac{1}{L}\sum_{t=1}^{L}
        \operatorname{clip}\!\left(A^{\mathrm{MOPD}}_t,-A_{\max},A_{\max}\right)
        \log\pi_{\theta}(z_t\mid h_t)
        \right],
    \end{aligned}
    \tag{14}
\end{equation}
where $\operatorname{sg}$ stops gradients through the advantage and $A_{\max}>0$ bounds its magnitude. Because the states come from the student while the signal comes from the appropriate expert, this stage directly repairs the behavioral interference left by merging without introducing an off-policy imitation gap. Minimizing Eq.~(14) yields the unified policy $\pi_{\mathrm{U}}$, which retains atomic financial reasoning and long-horizon agentic execution in a single checkpoint.

\section{Experiments}
\label{sec:experiments}

\subsection{Experimental Setup}

\paragraph{Models.} \ourcu{} (9B) and \ourag{} (27B) use Qwen3.5-9B and Qwen3.6-27B as their respective base models. We evaluate them against three groups of baselines. \textbf{Frontier models} include Gemini-3.5-Flash~\citep{gemini352026}, Claude Opus 4.8~\citep{claudeopus482026}, GPT-5.6-Sol~\citep{gpt562026}, GLM-5.2~\citep{glm522026}, MiniMax-M3~\citep{minimaxm32026}, Kimi-K2.7-Code~\citep{kimik272026}, DeepSeek-V4-Pro and DeepSeek-V4-Flash~\citep{deepseekv42026}, MiMo-V2.5-Pro~\citep{mimov252026}, and Qwen3.7-Plus~\citep{qwen372026}. \textbf{Open-source models} include Agents-A1-35B~\citep{agentsa12026}, Nex-N2-mini~\citep{nexn22026}, ASearcher-32B~\citep{asearcher2025}, OpenThinkerAgent-32B~\citep{openthinkeragent2026}, OpenResearcher-30B-A3B~\citep{openresearcher2026}, and Tongyi-DeepResearch-30B-A3B~\citep{tongyidr2025}. \textbf{Agent systems} include Codex with GPT-5.6~\citep{codex2026,gpt562026}, Cursor with Grok 4.5~\citep{cursorGrok452026}, and Cursor with Composer 2.5~\citep{cursorComposer252026}.

\paragraph{Benchmarks.} We evaluate two complementary capabilities. \textbf{Atomic financial reasoning} is measured by BizFinBench~\citep{bizfinbench2025}, FinanceBench~\citep{financebench2023}, and RFC-Bench Task 2~\citep{rfcbench2026}, covering business analysis, disclosure-grounded question answering, and counterfactual-misinformation classification. \textbf{Agentic execution} is measured by FinanceAgentBench v1.1 and v2~\citep{financeagentbench2025,financeagentbenchv22026} and FinSearchComp T2 and T3~\citep{finsearchcomp2025}, which require multi-step retrieval, calculation, and synthesis. Because no held-out FinanceAgentBench validation split is publicly available, and our inquiry to the benchmark maintainers regarding access to an additional split had not received a response at the time of evaluation, we use the 50 public v1.1 tasks and 27 public v2 tasks. FinSearchComp uses the original released test data. Detailed benchmark-specific verification protocols and judge prompts are provided in \Cref{app:benchmark-verification}.

\paragraph{Protocol.} All models are evaluated in a single turn on the three atomic financial-reasoning benchmarks. For FinanceAgentBench v1.1 and v2, we use the corresponding official evaluation harnesses.\footnote{\url{https://github.com/vals-ai/finance-agent} (v1.1) and \url{https://github.com/vals-ai/finance-agent-v2} (v2).} For FinSearchComp, we evaluate \ourcu{}, \ourag{}, and the frontier models with MintHarness. Among the open-source baselines, Tongyi DeepResearch\footnote{\url{https://github.com/Alibaba-NLP/DeepResearch}.}, ASearcher\footnote{\url{https://github.com/inclusionAI/ASearcher}.} and OpenResearcher\footnote{\url{https://github.com/TIGER-AI-Lab/OpenResearcher}.} retain their official inference harnesses, while the remaining open-source models are evaluated with MintHarness. We set the temperature to 0, repetition penalty to 1.1 and a 128K-token context per task. Evaluation prmopts/scripts are detailed in \Cref{app:benchmark-verification}. FinanceAgentBench v2 is run three times; \Cref{tab:financial-benchmarks} reports the mean and across-run variance.

\newcommand{\scorevar}[2]{%
    \ensuremath{#1_{\scriptscriptstyle\pm #2}}%
}

\begin{table*}[!t]
    \centering
    \caption{\textbf{Performance comparison on financial benchmarks.} All scores are percentages. FinanceAgentBench v2 reports the mean of three runs; the subscripted $\pm$ value is the variance. Green cells, from darkest to lightest, denote the best, second-best, and third-best result in each column.}
    \label{tab:financial-benchmarks}
    \vspace{-0.3em}
    \setlength{\tabcolsep}{8.5pt}
    \renewcommand{\arraystretch}{1.15}

    \resizebox{\textwidth}{!}{
    \begin{tabular}{lccccccc}
        \toprule

        \multirow{2}{*}{\textbf{Model}}
        & \multirow{2}{*}{\makecell{\textbf{BizFin-}\\\textbf{Bench}}}
        & \multirow{2}{*}{\makecell{\textbf{Finance}\\\textbf{Bench}}}
        & \multirow{2}{*}{\makecell{\textbf{RFC}\\\textbf{Bench}}}
        & \multicolumn{2}{c}{\textbf{FinanceAgentBench}}
        & \multicolumn{2}{c}{\textbf{FinSearchComp}}
        \\

        \cmidrule(lr){5-6}
        \cmidrule(lr){7-8}

        &
        &
        &
        & \textbf{v1.1}
        & \textbf{v2}
        & \textbf{T2}
        & \textbf{T3}
        \\

        \midrule
        \multicolumn{8}{l}{\textit{Frontier models}}
        \\

        Gemini-3.5-Flash
        & 50.71
        & 85.33
        & 92.33
        & 62.00
        & \scorevar{25.93}{0.09}
        & 36.53
        & 44.19
        \\

        Claude Opus 4.8
        & 49.71
        & \cellcolor{secondcell}90.00
        & \cellcolor{thirdcell}95.33
        & 66.00
        & \cellcolor{thirdcell}\scorevar{55.56}{0.00}
        & 45.66
        & \cellcolor{thirdcell}51.16
        \\

        GPT-5.6-Sol
        & \cellcolor{thirdcell}51.57
        & 86.67
        & 94.67
        & 66.00
        & \cellcolor{secondcell}\scorevar{56.79}{0.12}
        & 41.10
        & 50.58
        \\

        GLM-5.2
        & 47.57
        & 88.00
        & 94.33
        & 58.00
        & \scorevar{51.85}{0.09}
        & 68.49
        & 34.88
        \\

        MiniMax-M3
        & 40.71
        & 86.00
        & 91.33
        & \cellcolor{thirdcell}70.00
        & \scorevar{38.27}{0.03}
        & 60.27
        & 30.23
        \\

        Kimi-K2.7-Code
        & 46.57
        & 82.00
        & 92.67
        & 62.00
        & \scorevar{43.21}{0.12}
        & 56.62
        & 25.00
        \\

        DeepSeek-V4-Pro
        & 44.00
        & 87.33
        & 91.33
        & 60.00
        & \scorevar{40.74}{0.09}
        & 60.73
        & 31.40
        \\

        DeepSeek-V4-Flash
        & 45.71
        & \cellcolor{thirdcell}88.67
        & 89.67
        & \cellcolor{secondcell}72.00
        & \scorevar{43.21}{0.03}
        & 59.82
        & 25.00
        \\

        MiMo-V2.5-Pro
        & 45.00
        & 87.33
        & 91.33
        & 66.00
        & \scorevar{37.04}{0.09}
        & 58.45
        & 26.16
        \\

        Qwen3.7-Plus
        & 48.43
        & 88.00
        & 93.33
        & 64.00
        & \scorevar{48.15}{0.09}
        & 62.10
        & 31.98
        \\

        \midrule
        \multicolumn{8}{l}{\textit{Open-source models}}
        \\

        Agents-A1-35B
        & 41.86
        & 84.00
        & 90.00
        & 42.00
        & \scorevar{11.11}{0.00}
        & 47.03
        & 11.63
        \\

        Nex-N2-mini
        & 27.43
        & 69.33
        & 83.33
        & 38.00
        & \scorevar{7.41}{0.09}
        & 57.08
        & 20.35
        \\

        ASearcher-32B
        & 28.14
        & 81.33
        & 84.00
        & 24.00
        & \scorevar{3.70}{0.00}
        & 42.92
        & 4.07
        \\

        OpenThinkerAgent-32B
        & 25.00
        & 66.67
        & 71.00
        & 34.00
        & \scorevar{0.00}{0.00}
        & 36.07
        & 14.53
        \\

        OpenResearcher-30B-A3B
        & 10.71
        & 50.00
        & 22.00
        & 42.00
        & \scorevar{0.00}{0.00}
        & 53.88
        & 16.86
        \\

        Tongyi-DeepResearch-30B-A3B
        & 19.57
        & 42.67
        & 57.67
        & 58.00
        & \scorevar{9.88}{0.03}
        & 72.15
        & 18.02
        \\

        \midrule
        \multicolumn{8}{l}{\textit{Agent systems}}
        \\

        Codex (GPT-5.6)
        & 46.14
        & 85.33
        & 86.33
        & 66.00
        & \scorevar{54.32}{0.03}
        & \cellcolor{thirdcell}80.37
        & \cellcolor{bestcell}54.07
        \\

        Cursor (Grok 4.5)
        & 48.29
        & 88.00
        & 63.33
        & 66.00
        & \scorevar{53.09}{0.12}
        & \cellcolor{secondcell}81.74
        & \cellcolor{secondcell}52.33
        \\

        Cursor (Composer 2.5)
        & 47.00
        & 84.67
        & 69.67
        & \cellcolor{secondcell}72.00
        & \scorevar{49.38}{0.03}
        & 77.63
        & \cellcolor{thirdcell}51.16
        \\

        \midrule
        \multicolumn{8}{l}{\textit{Ours}}
        \\

        \textbf{Mint-Cu}
        & \cellcolor{secondcell}53.86
        & \cellcolor{secondcell}90.00
        & \cellcolor{secondcell}96.67
        & 68.00
        & \scorevar{41.98}{0.21}
        & 69.86
        & 34.88
        \\

        \textbf{Mint-Ag}
        & \cellcolor{bestcell}55.71
        & \cellcolor{bestcell}91.33
        & \cellcolor{bestcell}98.33
        & \cellcolor{bestcell}76.00
        & \cellcolor{bestcell}\scorevar{60.49}{0.03}
        & \cellcolor{bestcell}89.04
        & \cellcolor{bestcell}54.07
        \\

        \bottomrule
    \end{tabular}
    }

\end{table*}

\subsection{Overall Performance}
\label{sec:overall-performance}

\paragraph{Atomic Reasoning.} \Cref{tab:financial-benchmarks} shows that the two Mint models consistently occupy the leading tier on bounded financial tasks. \ourag{} achieves the best BizFinBench score at 55.71, exceeding the strongest non-Mint result by 4.14 points, and also leads FinanceBench at 91.33 and RFC-Bench at 98.33, improving over the strongest external baselines by 1.33 and 3.00 points, respectively. These benchmarks emphasize complementary forms of financial reasoning: BizFinBench stresses business analysis across heterogeneous task types, FinanceBench requires precise question answering over financial disclosures, and RFC-Bench tests whether the model can identify source-grounded financial misinformation. Strong performance across all three therefore reflects more than isolated numerical accuracy; it requires the model to preserve financial semantics, reporting scope, and evidential consistency across different answer formats. The joint strength across analysis, disclosure QA, and claim verification indicates that scaling the agentic model does not trade away precise financial reasoning.

\paragraph{Agentic Execution.} The advantage is larger when tasks require open-ended research. \ourag{} reaches 76.00 on FinanceAgentBench v1.1, 60.49 on v2, and 89.04 on FinSearchComp T2, outperforming the best external scores on these benchmarks---72.00 from DeepSeek-V4-Flash and Cursor (Composer 2.5), 56.79 from GPT-5.6-Sol, and 81.74 from Cursor (Grok 4.5)---by 4.00, 3.70, and 7.30 points. These tasks require the model to move beyond answering from a supplied context: it must locate relevant sources, maintain the temporal and financial scope of the question, carry intermediate values across multiple steps, and terminate with a complete answer. The consistent gains across FinanceAgentBench and FinSearchComp suggest that the resulting execution policy transfers across distinct research objectives rather than specializing to a single task format. On the more demanding FinSearchComp T3 track, \ourag{} matches the top score of 54.07, while the compact 9B \ourcu{} obtains 69.86 on T2, showing that substantial long-horizon execution capability is already retained at the smaller scale.

\paragraph{Capability Balance.} Taken together, the results highlight the complementary roles of the two Mint models. \ourcu{} provides a compact operating point that retains strong bounded reasoning and competitive research execution, whereas \ourag{} extends this foundation and improves performance across the full evaluation suite. More importantly, the gains on long-horizon tasks do not come at the expense of disclosure-grounded reasoning or verification. This balance supports the central design objective of \ourmethod: integrating reliable financial reasoning and sustained agentic execution within a single policy.

\begin{figure}[!tbp]
    \centering
    \begin{subfigure}[t]{0.485\linewidth}
        \centering
        \includegraphics[width=\linewidth]{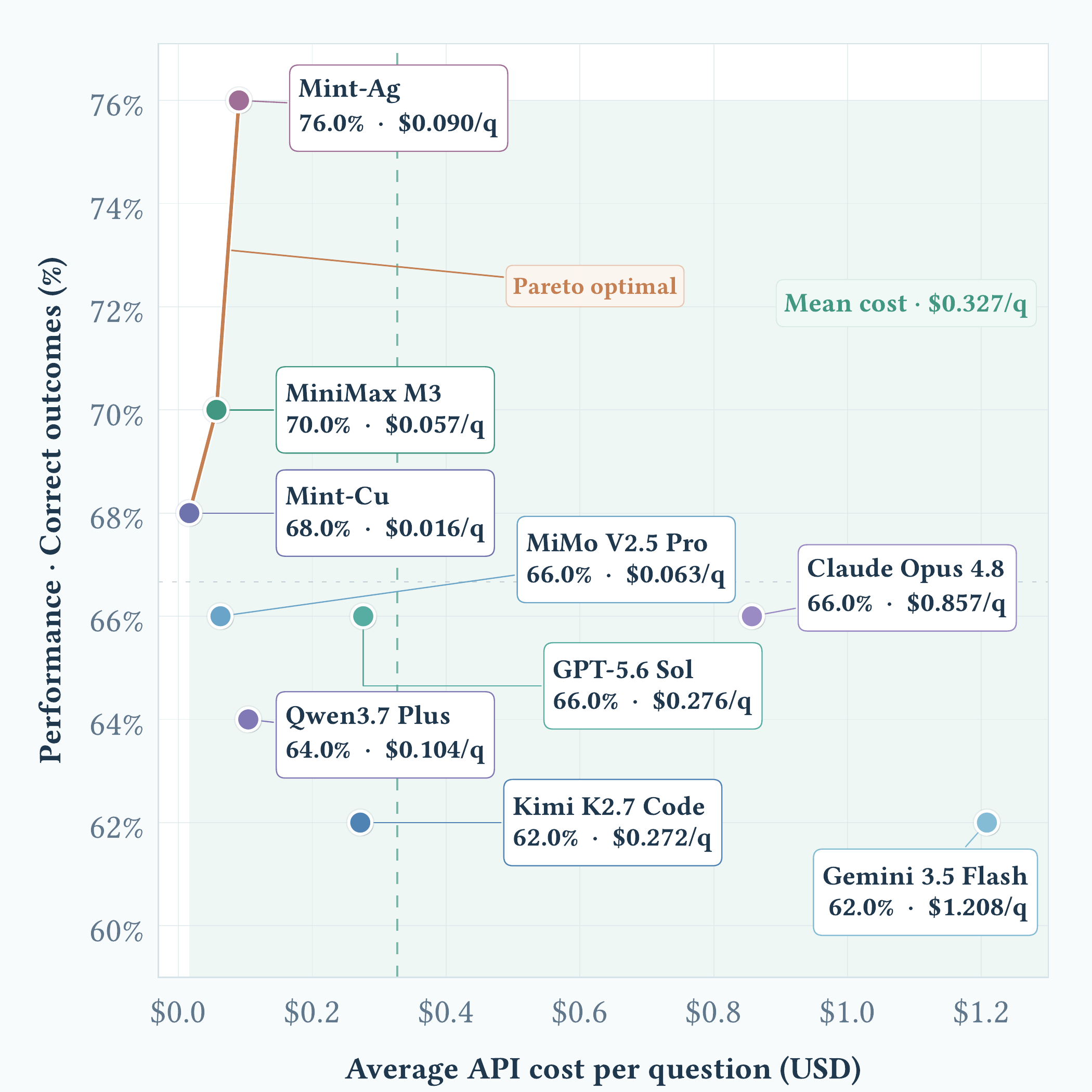}
        \caption{FinanceAgentBench v1.1.}
        \label{fig:cost-v11}
    \end{subfigure}
    \hfill
    \begin{subfigure}[t]{0.485\linewidth}
        \centering
        \includegraphics[width=\linewidth]{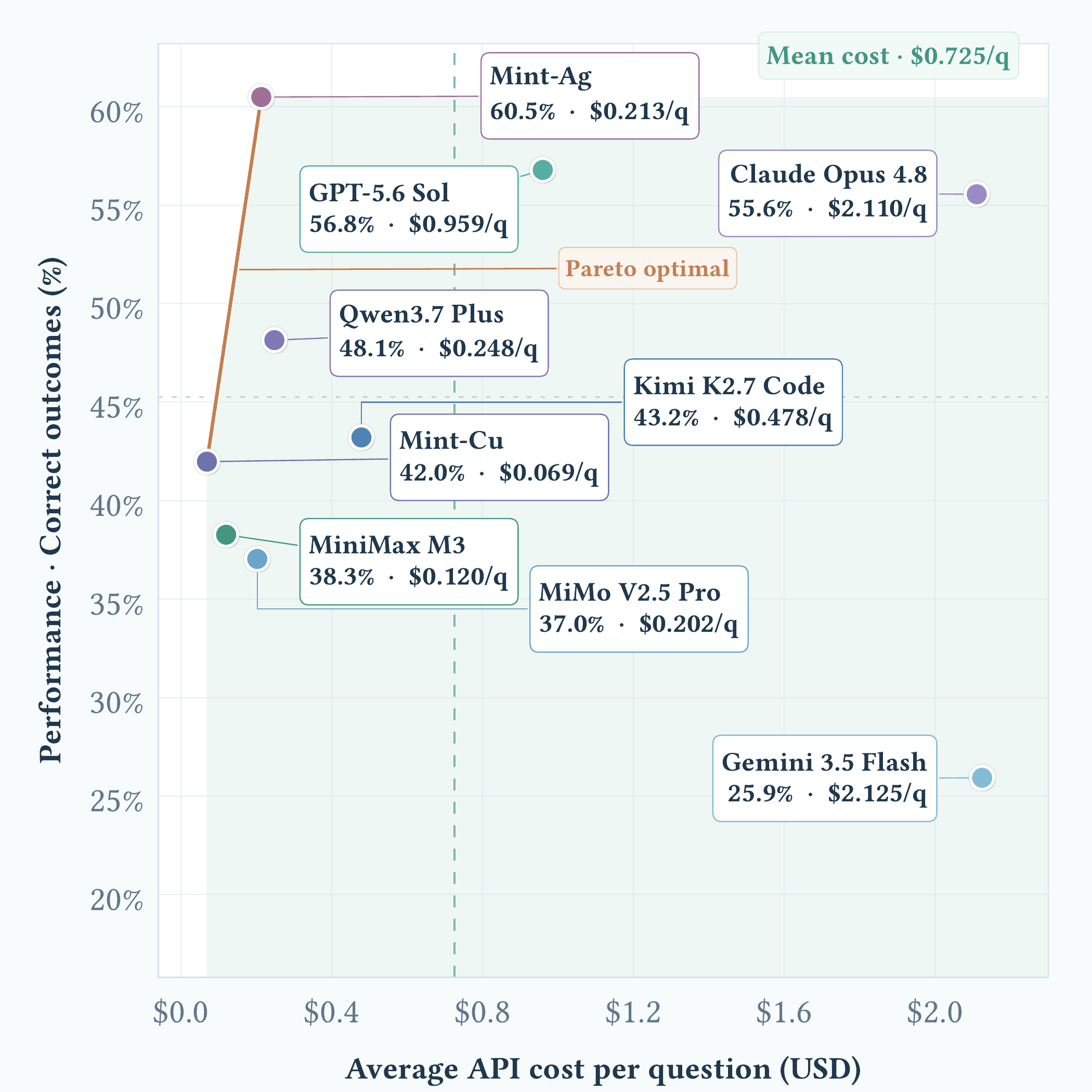}
        \caption{FinanceAgentBench v2.}
        \label{fig:cost-v2}
    \end{subfigure}
    \caption{\textbf{Cost--performance trade-offs on FinanceAgentBench.}
    Correct outcomes are plotted against average API cost per question. The
    dashed vertical line marks the mean cost of the compared models, and the
    highlighted curve traces the empirical Pareto frontier.}
    \label{fig:cost-performance}
    \vspace{-0.2em}
\end{figure}

\subsection{Cost Analysis}

\paragraph{Pareto Efficiency.}
\Cref{fig:cost-performance} jointly compares task accuracy and average API cost per question. Both Mint models lie on the empirical Pareto frontier in both FinanceAgentBench versions: no evaluated alternative is simultaneously more accurate and less expensive. On v1.1, \ourcu{} anchors the low-cost region with 68.0\% accuracy at \$0.016 per task, while \ourag{} reaches the overall-best 76.0\% at \$0.090---72.5\% below the \$0.327 comparison mean. On v2, \ourcu{} is again the least expensive point at \$0.069, and \ourag{} attains the highest accuracy of 60.49\% at \$0.213. Relative to the strongest external model, GPT-5.6-Sol, \ourag{} improves accuracy by 3.70 points while reducing cost by 77.8\% (\$0.213 versus \$0.959). Thus, the advantage is not low cost in isolation: the two models trace complementary low-cost and high-performance operating points on the Pareto frontier. For these locally hosted checkpoints, we estimate API-equivalent cost by applying the prices of their respective base models (Qwen3.5-9B for \ourcu{} and Qwen3.6-27B for \ourag{}) to measured token usage.

\subsection{Model Analysis}
\label{sec:training-analysis}

\begin{figure}[!t]
    \centering
    \vspace{-0.1em}
    \includegraphics[width=0.94\linewidth]{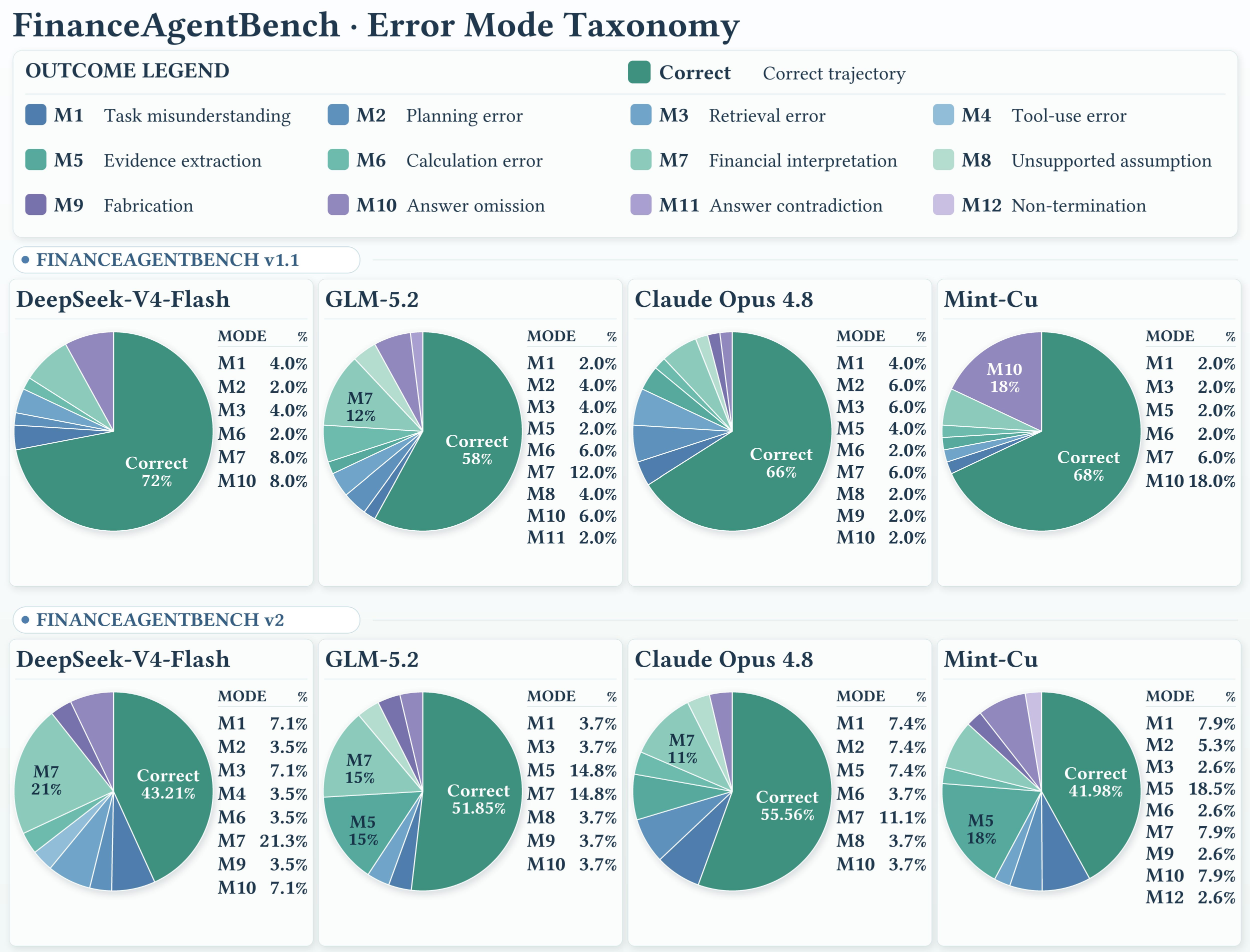}
    \vspace{-0.2em}
    \caption{\textbf{Failure taxonomy on FinanceAgentBench.} Correct outcomes and primary failure modes are shown for representative frontier models and \ourcu{} on v1.1 and v2.}
    \label{fig:failure-modes}
    \vspace{-0.1em}
\end{figure}

\paragraph{Failure Modes.}
To construct the failure taxonomy, PhD researchers with backgrounds in both finance and AI manually review every trajectory and assign its earliest consequential error. \Cref{fig:failure-modes} summarizes these annotations. On v1.1, \ourcu{} solves 68\% of the tasks; its largest residual failure mode is answer omission (M10, 18\%), whereas task misunderstanding, retrieval, evidence extraction, and calculation each account for only 2\%. On v2, accuracy falls to 41.98\% and the dominant error shifts to evidence extraction (M5, 18.5\%). Task misunderstanding, financial interpretation, and answer omission each contribute a further 7.9\%, while retrieval, calculation, fabrication, and non-termination remain individually small. This shift suggests that the harder benchmark is constrained less by accessing sources than by converting retrieved material into complete, correctly scoped financial evidence.

\paragraph{Effect of MOPD.}

\begin{table}[!tbp]
    \centering
    \small
    \setlength{\tabcolsep}{6pt}
    \caption{Effect of expert integration for the 9B model. All scores are percentages; \textemdash{} denotes a specialist not evaluated outside its target domain.}
    \label{tab:mopd-ablation}
    \begin{tabular}{lcccc}
        \toprule
        \textbf{Checkpoint} & \makecell{\textbf{Finance}\\\textbf{Bench}} & \makecell{\textbf{BizFin}\\\textbf{Bench}} & \makecell{\textbf{FAB}\\\textbf{v1.1}} & \makecell{\textbf{FinSearch}\\\textbf{T2}} \\
        \midrule
        Reasoning expert ($\pi_{\mathrm{R}}$) & 88.00& \textbf{53.86} & \textemdash & \textemdash \\
        Execution expert ($\pi_{\mathrm{A}}$) & \textemdash & \textemdash & 66.21& \textbf{72.15} \\
        TIES merge ($\pi_{\mathrm{M}}$) & 88.67 & 50.00 & 64.00 & 66.67 \\
        MOPD ($\pi_{\mathrm{U}}$) & \textbf{90.00}& \textbf{53.86} & \textbf{68.00}& 69.86 \\
        \bottomrule
    \end{tabular}
\end{table}

\Cref{tab:mopd-ablation} isolates the integration stage, with each specialist evaluated only in its target domain. TIES produces uneven transfer across the four benchmarks, whereas MOPD improves the merged checkpoint by 1.33, 3.86, 4.00, and 3.19 points, respectively. The resulting scores align with the \ourcu{} results in \Cref{tab:financial-benchmarks}, showing that MOPD restores performance across both capability groups.

\begin{figure}[!t]
    \centering
    \includegraphics[width=\linewidth]{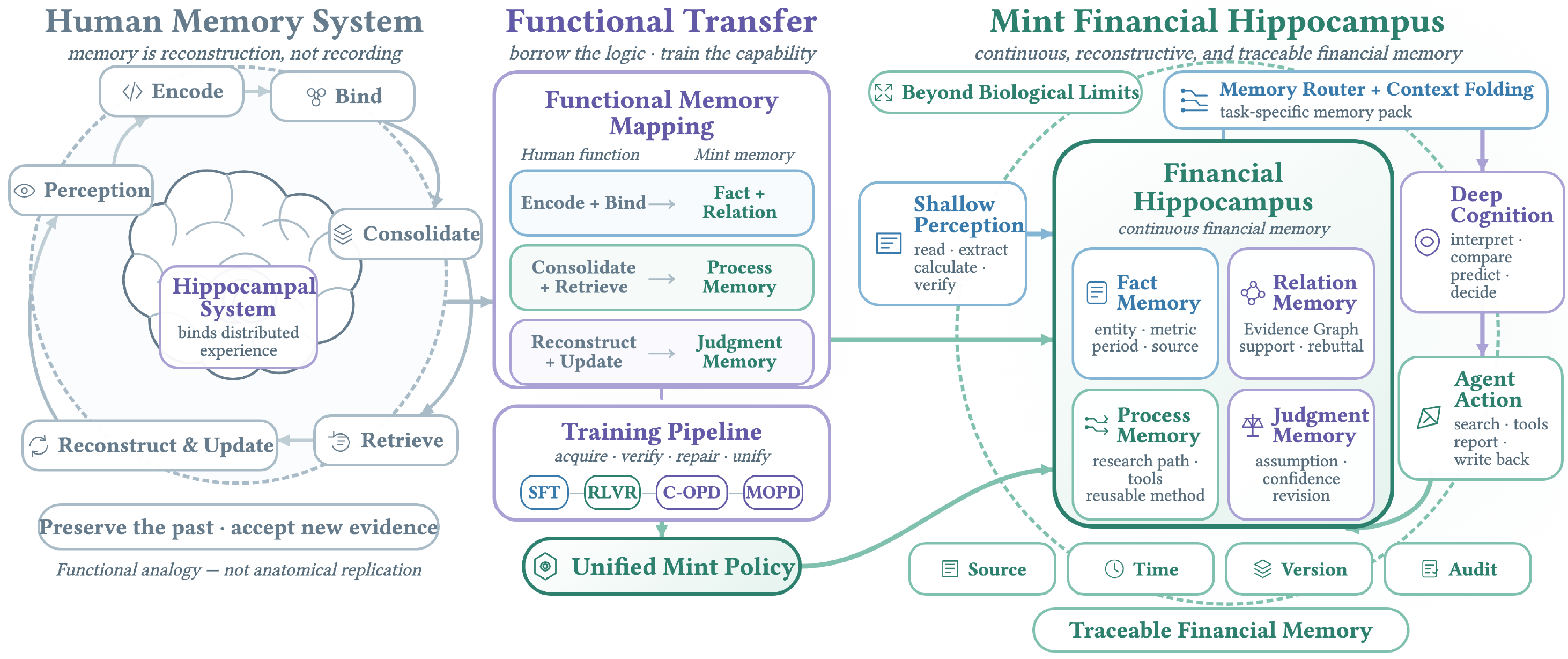}
    \caption{The mechanism analysis of \ourmethod from the cognitive architecture view.}
    \label{fig:brain}
\end{figure}

\paragraph{Mechanism Analysis} Read together with the failure taxonomy, this specialization-and-integration pattern suggests a structured cognitive architecture. MintHarness first turns observations into an external perceptual memory: the Evidence Ledger preserves what was observed, from which source, and at what time, mitigating memory decay and source confusion over long trajectories. Compact working memory then selects from this durable record only the current goal, unresolved subquestions, and evidence needed for the next decision, limiting overload and attentional drift without duplicating the ledger. Above these memory layers, the reasoning expert supplies a deliberative core for finance-sensitive interpretation and calculation, while the execution expert supplies an executive core for search, tool use, progress monitoring, and termination. MOPD consolidates the two into one policy, retaining their division of labor while reducing skill interference and the coordination cost of switching between separate specialists.

This organization connects the architectural analogy to observed behavior. Separating durable perception from active manipulation helps protect precise reasoning from context noise, while separating deliberation from executive control keeps long-horizon search tied to an answerable objective. The dominant answer-omission failures on v1.1 resemble a breakdown in executive closure: useful intermediate work is not converted into a final response. On v2, the shift toward evidence-extraction failures moves the bottleneck earlier, to the perceptual-memory--working-memory interface: a source may be reached, yet the financially relevant fact is not selected and encoded with the correct scope. The gains after MOPD indicate that the two cores can be unified without sacrificing their complementary strengths; the remaining taxonomy points instead to selective attention over evidence as the next capability to improve.

\subsection{Case Study}
\label{sec:case-study}

We inspect two successful FinanceAgent trajectories covering temporal disambiguation, long-horizon recomputation, and multi-source reconciliation. The cases illustrate how the Evidence Ledger supports these research patterns. An additional trajectory illustrating retrieval recovery and replayable calculation is provided in \Cref{app:additional-case-study}.

\begin{figure}[!tbp]
    \centering
    \includegraphics[width=\linewidth,height=1\textheight,keepaspectratio]{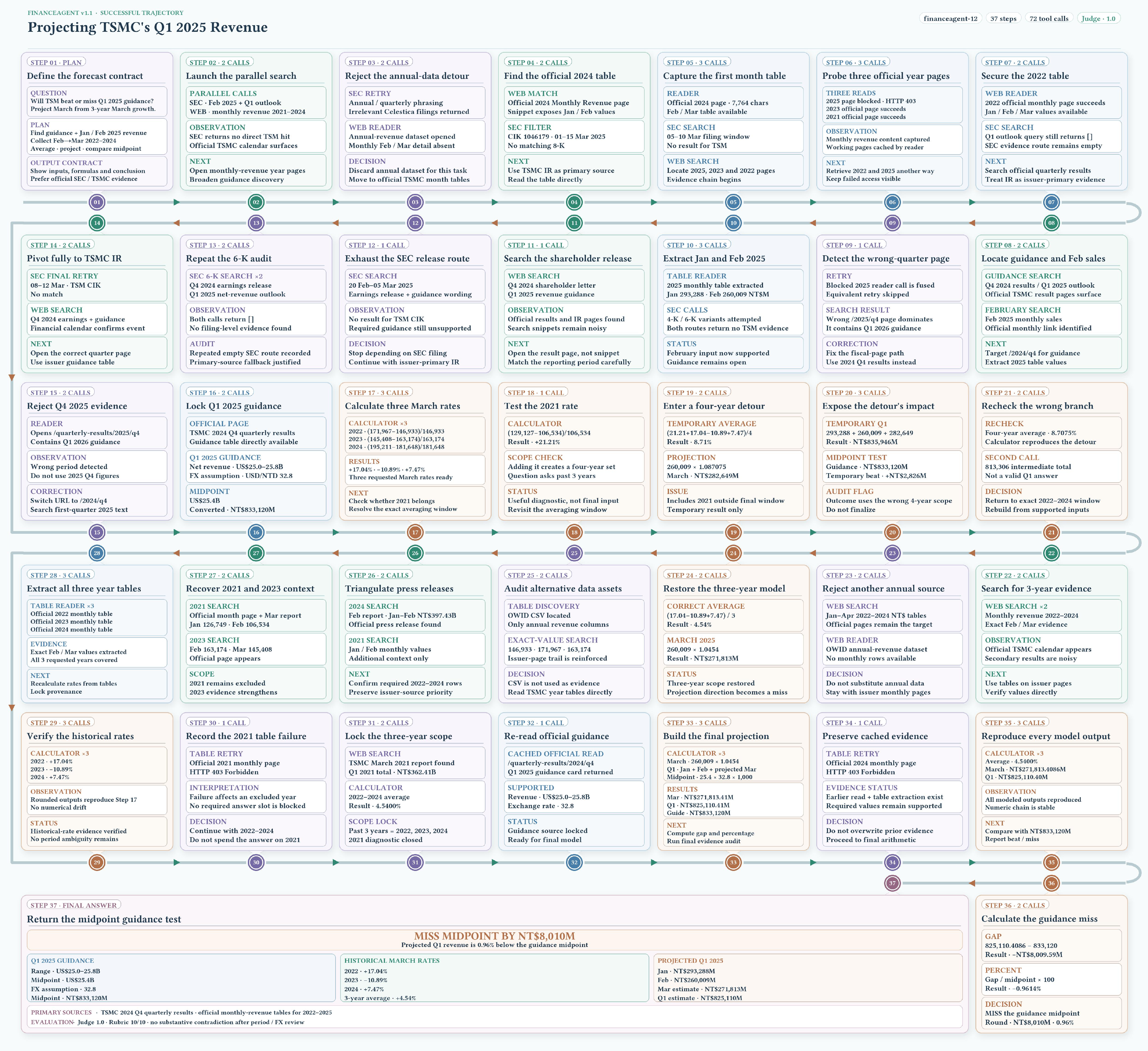}
    \caption{\textbf{TSMC forecast.} Persistent evidence and compact working state carry valid guidance and monthly revenue records through 37 steps of source failures, scope corrections, and recomputation.}
    \label{fig:case-tsm}
\end{figure}

\paragraph{TSMC Forecast.}
The 37-step trajectory in \Cref{fig:case-tsm} stresses persistent state over a substantially longer investigation. It encounters blocked filings, wrong-quarter pages, annual datasets, and an initially incorrect four-year averaging window, but retains the valid Q1 2025 guidance and the monthly revenue records already recovered. After restoring the intended 2022--2024 window, the agent recomputes the historical March growth rate, projects March 2025 revenue, and estimates quarterly revenue at NT\$825{,}110 million. Because the guidance midpoint and exchange-rate conversion remain locked in the ledger, the final NT\$8{,}010 million miss can be reproduced without trusting the intermediate detours.

\begin{figure}[!tbp]
    \centering
    \includegraphics[width=\linewidth,height=0.82\textheight,keepaspectratio]{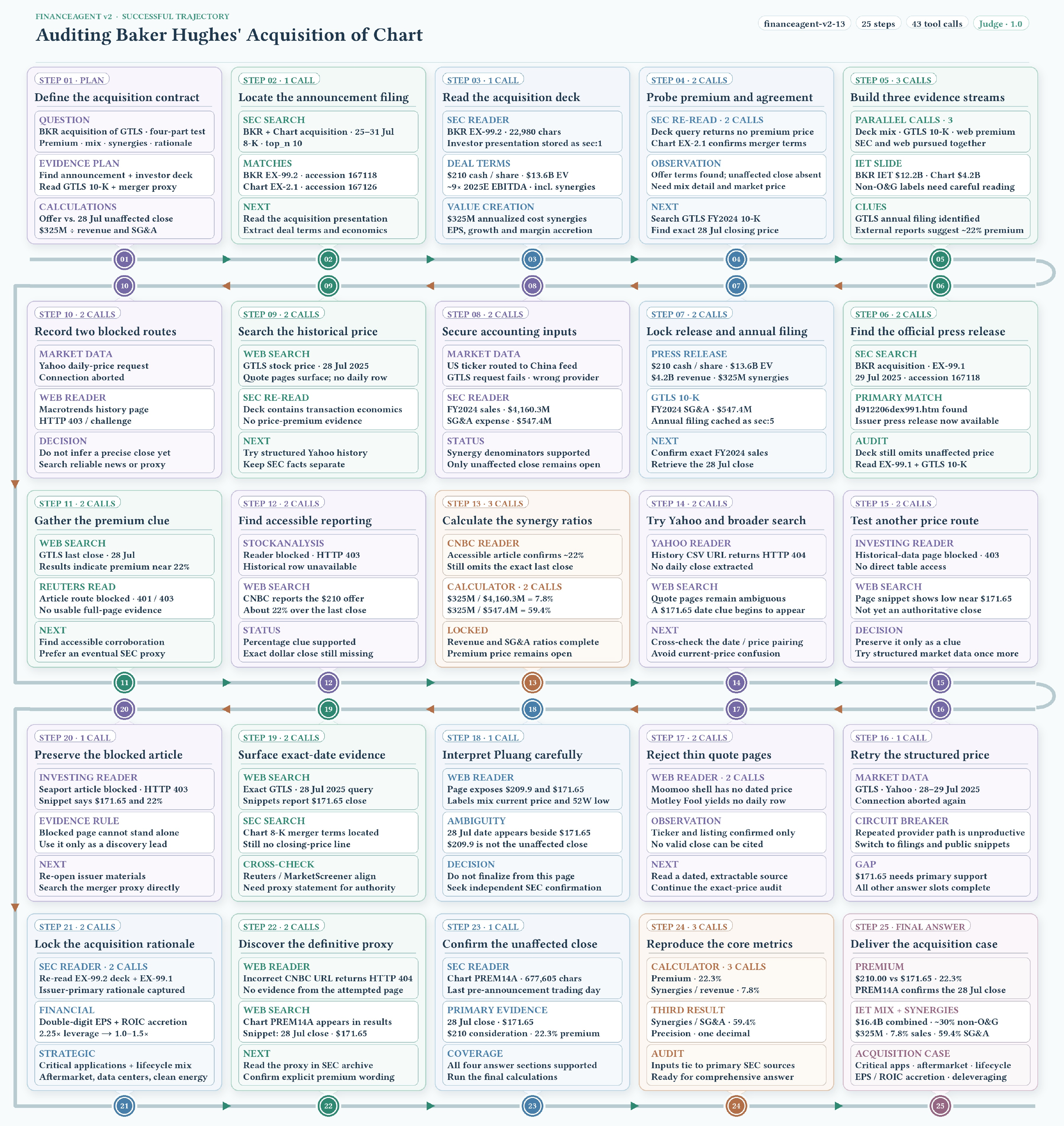}
    \caption{\textbf{BKR acquisition.} Multiple evidence streams are reconciled only after the unaffected closing price is confirmed in the definitive proxy, enabling reproducible premium and synergy calculations.}
    \label{fig:case-bkr}
\end{figure}

\paragraph{BKR Acquisition.}
\Cref{fig:case-bkr} combines deal terms, accounting inputs, market data, and acquisition rationale. Most fields are recovered early from the announcement materials and GTLS filing, but the unaffected closing price remains unresolved across several blocked or ambiguous sources. The agent keeps those pages as discovery leads rather than treating them as evidence, then locks the exact \$171.65 close from the definitive proxy. With that missing input resolved, the ledger supports a 22.3\% premium and reproduces the \$325 million synergy ratios against both revenue and SG\&A, while preserving the primary source behind every reported quantity.

\section{Conclusion}
\label{sec:conclusion}

We presented \ourmethod, a full-stack approach to building finance-native agentic foundation models around a single notion of financial correctness. Its data engine jointly constructs atomic financial capabilities and long-horizon research trajectories from provenance-linked sources; MintHarness turns open-ended interaction into a recoverable execution process through the evidence ledger and context management; and the training stack combines supervised fine-tuning, critical-step OPD, and RLVR before integrating reasoning and execution experts with TIES and multi-teacher OPD. Together, these components connect what the model learns, how it acts, and how its conclusions are verified, rather than treating domain knowledge, tool use, and auditability as separate concerns.

This design is realized in two complementary models. \ourcu{} shows that a compact 9B model can sustain strong financial research performance at favorable inference cost, while \ourag{} extends the same foundation to more demanding reasoning and execution. Across seven benchmarks, \ourag{} reaches 98.33 on RFC-Bench, 76.00 and 60.49 on FinanceAgentBench v1.1 and v2, and 89.04 on FinSearchComp T2. The trajectory studies further show that these gains are not limited to final-answer accuracy: the models can preserve evidence, calculations, and intermediate decisions across multi-step investigations. 

The broader lesson is that financial intelligence cannot be reduced to producing a plausible answer. A trustworthy financial agent must be able to recover the evidence it relied on, respect the temporal boundary of the question, expose the transformations that connect facts to claims, and revise its conclusion when any link fails. From this perspective, auditability is not a reporting layer added after inference; it is a shared substrate for data construction, agent execution, and policy learning. \ourmethod{} is a step toward that foundation: financial agents whose competence is measured not only by what they conclude, but also by whether each conclusion can be traced and tested. In high-stakes research, intelligence becomes useful at scale only when it leaves such a trail.




\section{Contributors}
\label{sec:contri}

\paragraph{Project Lead} Gavin Zhang

\paragraph{Core Contributors}
Yaze Geng, Lei Tang, Yaoyang Yi, Zonghan Wu

\paragraph{Contributors}
Yifan Hu

\paragraph{Corresponding authors}
Kun Wang, Qingsong Wen,  Yilei Shao


\bibliographystyle{assets/plainnat}
\bibliography{references}


\clearpage
\appendix
\section{Benchmark Verification Protocols}
\label{app:benchmark-verification}

This appendix documents how model outputs are converted into benchmark scores. We preserve each benchmark's released evaluator, reference answers, or per-example rubrics rather than introducing a shared judge across datasets. The prompts below omit only lengthy in-prompt demonstrations where noted; the corresponding released prompts are used unchanged during evaluation.

\subsection{BizFinBench}

BizFinBench is scored with the official BizFinBench v2 evaluation scripts. The evaluator selects the released comparison function associated with each sample's \texttt{task\_type}, including anomaly tracing, event-logic reasoning, financial data description, multi-turn perception, quantitative computation, report analysis, stock-price prediction, and sentiment analysis. Our interface extracts the model-visible final answer and performs only the formatting required by the relevant official script, such as wrapping a scalar answer, JSON array, boxed value, or interval. The normalized prediction and the original sample metadata are then passed to the official evaluator, and a returned score of at least 1.0 is counted as correct. This protocol is deterministic and uses no LLM judge prompt.

\subsection{FinanceBench}

FinanceBench is evaluated using its released reference answers and official answer-matching procedure. The evaluator receives the model's final answer after answer extraction; no additional task-solving context or external evidence is supplied by the evaluator. We report the resulting question-level accuracy.

\subsection{FinSearchComp T2 and T3}

For both FinSearchComp T2 and T3, we use the official \texttt{judge\_system\_prompt} and \texttt{judge\_prompt\_template} stored with each released example. The two tracks use the same judge template. The judge compares the extracted student answer with the reference answer and any accompanying scoring points, then returns a binary \texttt{answer\_score}. Multi-part answers receive credit only when all required elements are correct; numerically equivalent forms are accepted, and an omitted unit is accepted when the question already specifies that unit. Empty or error-valued answers receive zero. The official prompt also contains four demonstrations, which are retained during evaluation but omitted here for space.

\paragraph{Chinese system prompt used by the evaluator.}
\begin{promptbox}
\begin{CJK*}{UTF8}{gbsn}
\small
你是一个金融问题答案的判断和评分者。给定一道题目、标准答案（可能附有评分要点）和学生答案，先识别学生答案中的最终答案，再与标准答案及评分要点比较。只有全部必要要点均正确时得 1 分，否则得 0 分；空答案或异常报错得 0 分。数值的等价格式视为一致；当题目已经指定单位时，学生答案可以省略单位。不要自行回答或解决题目。输出简短的评分依据，并以 JSON 字段 \texttt{answer\_score} 返回 0 或 1。
\end{CJK*}
\end{promptbox}

\paragraph{English rendering.}
\begin{promptbox}
\small
You are a judge and scorer of answers to financial questions. Given a question, a reference answer that may include scoring points, and a student answer, first identify the student's final answer and then compare it with the reference answer and scoring points. Award 1 only if every required element is correct; otherwise award 0. Empty or error-valued answers receive 0. Numerically equivalent formats are treated as equal, and a unit may be omitted when it is specified by the question. Do not solve the question yourself. Return a brief rationale and a JSON object whose \texttt{answer\_score} is 0 or 1.
\end{promptbox}

\paragraph{Official user template.}
\begin{promptbox}
\begin{CJK*}{UTF8}{gbsn}
\small
\texttt{<题目>：} \{prompt\}\\
\texttt{<标准答案>：} \{response\_reference\}\\
\texttt{<学生答案>：} \{response\}
\end{CJK*}
\end{promptbox}

\subsection{FinanceAgentBench v1.1 and v2}

FinanceAgentBench is scored from the correctness rubrics released with each example. An LLM judge receives the question, expert reference answer, model prediction, and ordered list of correctness criteria. It determines whether the prediction satisfies each criterion and separately checks whether any confident factual claim directly contradicts the expert reference. Formatting, ordering, and writing style do not affect a match; reasonable rounding is accepted when units and orders of magnitude agree. A sample is counted as correct only when every correctness criterion is satisfied and no contradiction is detected. If an example contains no correctness criterion, the evaluator falls back to exact, substring, and numeric-overlap answer matching. The same protocol is used for v1.1 and v2.

\paragraph{Judge system prompt.}
\begin{promptbox}
\small
You are a strict but fair judge for the Finance Agent Benchmark. You will be given a financial question, an expert reference answer, the agent's final answer, and a list of rubric criteria. Decide, for each criterion, whether the final answer satisfies it, and separately decide whether the answer contradicts the expert reference on any specific fact. Do not solve the question or use outside knowledge. Return only valid JSON containing the ordered criterion-level decisions, a contradiction decision, and a short overall rationale.
\end{promptbox}

\paragraph{Judge user template.}
\begin{promptbox}
\small
\texttt{[Question]} \{question\}\\
\texttt{[Expert reference answer]} \{reference\}\\
\texttt{[Agent's final answer]} \{prediction\}\\
\texttt{[Rubric criteria]} \{rubric\_block\}
\end{promptbox}

\paragraph{Illustrative rubric example.}
The following synthetic example shows the complete criterion-level decision process; it is not an item from the benchmark test set.

\begin{promptbox}
\small
\textbf{[Question]}\\
Using the company's FY2024 filing, report its revenue, year-over-year revenue growth, and operating margin.\\[0.4em]

\textbf{[Expert reference answer]}\\
FY2024 revenue was \$10.2 billion, representing 8.5\% year-over-year growth, and operating margin was 14.3\%.\\[0.4em]

\textbf{[Agent's final answer]}\\
The company reported FY2024 revenue of \$10.2 billion, up 8.5\% year over year, with an operating margin of 14.0\%.\\[0.4em]

\textbf{[Rubric criteria]}\\
0. States that FY2024 revenue was \$10.2 billion.\\
1. States that FY2024 revenue increased 8.5\% year over year.\\
2. States that FY2024 operating margin was 14.3\%.\\[0.4em]

\textbf{[Judge output]}\\
\texttt{\{}\\
\quad\texttt{"criteria\_results": [}\\
\qquad\texttt{\{"index": 0, "matched": true, "rationale": "Revenue matches."\},}\\
\qquad\texttt{\{"index": 1, "matched": true, "rationale": "Growth rate matches."\},}\\
\qquad\texttt{\{"index": 2, "matched": false, "rationale": "14.0\% does not match 14.3\%."\}}\\
\quad\texttt{],}\\
\quad\texttt{"contradiction": \{"detected": true, "rationale": "The operating-margin figure conflicts with the reference."\},}\\
\quad\texttt{"overall\_rationale": "Two of three criteria are satisfied, but the operating margin is incorrect."}\\
\texttt{\}}
\end{promptbox}

Here, the criterion-level rubric score is $2/3$, but the benchmark's binary sample score is 0 because not all correctness criteria are satisfied and the prediction contains a factual contradiction.

\subsection{RFC-Bench Task 2}

We evaluate only RFC-Bench Task 2, which asks the model to classify the principal manipulation in a source-grounded counterfactual financial paragraph. This benchmark uses no LLM judge. The evaluator extracts one of four labels---\texttt{numerical}, \texttt{flipping}, \texttt{sentiment}, or \texttt{causal}---from the model-visible final answer, normalizes a small set of label aliases, and performs exact matching against the reference perturbation type. Ambiguous or invalid outputs are counted as incorrect. The reported RFC-Bench result is therefore Task-2 classification accuracy rather than an aggregate over the full RFC-Bench suite.

\paragraph{Task and output prompt.}
\begin{promptbox}
\small
RFC-BENCH Task 2: source-grounded counterfactual financial misinformation type classification. Given the source article or page content and a manipulated financial news paragraph, identify which manipulation type was applied. Use only the provided source page and manipulated paragraph; do not browse, retrieve, or use outside evidence. Choose the single label that best captures the main manipulation mechanism: \texttt{numerical} for altered quantitative information, \texttt{flipping} for reversed direction or polarity, \texttt{sentiment} for amplified or weakened evaluative tone, or \texttt{causal} for a distorted cause, driver, attribution, or consequence. Output exactly one of these four labels and no additional text.
\end{promptbox}

\section{Additional Case Study}
\label{app:additional-case-study}

The following case complements the TSMC forecast and BKR acquisition trajectories discussed in \Cref{sec:case-study}. It focuses on retrieval recovery and the reuse of previously collected evidence during a multi-period financial calculation.

\begin{figure}[H]
    \centering
    \vspace{-0.3em}
    \includegraphics[width=\linewidth,height=0.82\textheight,keepaspectratio]{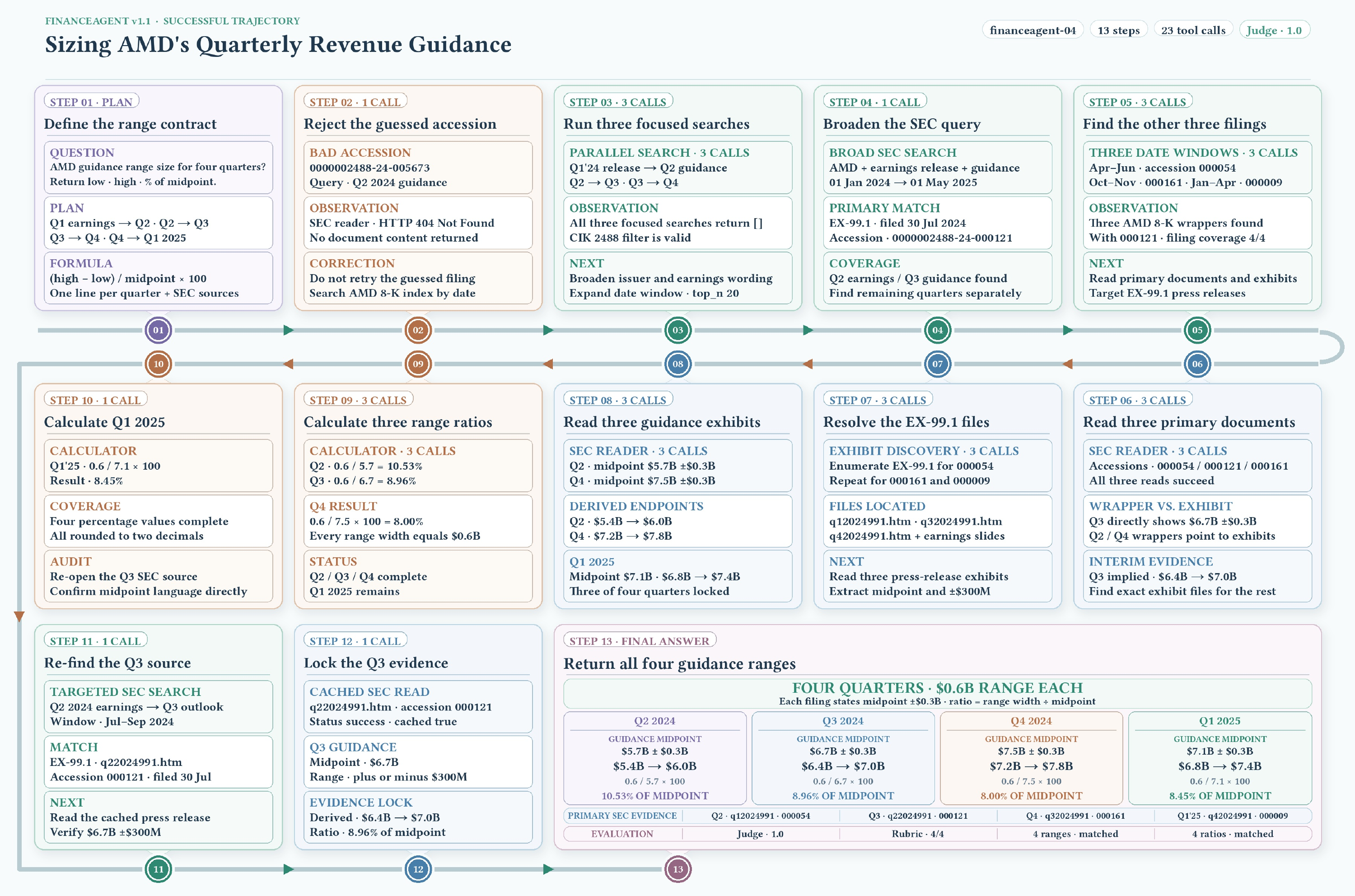}
    \caption{\textbf{AMD guidance.} A failed document lookup is replaced by broader SEC discovery, exhibit-level extraction, and a replayable calculation over four quarterly guidance records.}
    \label{fig:case-amd}
    \vspace{-0.4em}
\end{figure}

\paragraph{AMD Guidance.}
\Cref{fig:case-amd} follows a 13-step trajectory that sizes AMD's revenue-guidance ranges across four quarters. The initial accession guess fails, after which the agent broadens the SEC search, locates the relevant EX-99.1 exhibits, and records the midpoint and range endpoints for each quarter. The ledger allows the agent to reopen the Q3 source without repeating the full search. The final answer is supported by four primary filings and reports range widths of 10.53\%, 8.96\%, 8.00\%, and 8.45\% of midpoint, respectively.

\end{document}